\documentclass[10pt,twocolumn,letterpaper]{article}

\usepackage{verbatim}
\usepackage{booktabs}       
\usepackage{style}
\usepackage[pagenumbers]{cvpr} 
\usepackage{float}
\usepackage{multirow}
\usepackage{threeparttable}
\usepackage[accsupp]{axessibility} 

\usepackage[dvipsnames]{xcolor}
\definecolor{cvprblue}{rgb}{0.21,0.49,0.74}
\usepackage[pagebackref,breaklinks,colorlinks,citecolor=cvprblue]{hyperref}

\def\paperID{11} 
\def\confName{CVPR Workshop}
\def\confYear{2024}

\title{Hinge-Wasserstein: Estimating Multimodal Aleatoric Uncertainty\\in Regression Tasks 
}

\author{%
    Ziliang Xiong$^{1}$, Arvi Jonnarth$^{1,3}$, Abdelrahman Eldesokey$^{2}$,\\ Joakim Johnander$^{4}$, Bastian Wandt$^{1}$, Per-Erik Forssén$^{1}$ \\
    ${}^1$Computer Vision Laboratory, Department of Electrical Engineering, LiU, Sweden \\
    ${}^2$Visual Computing Center, KAUST, Saudi Arabia\\
    ${}^3$Husqvarna Group, Huskvarna, Sweden, ${}^4$Zenseact, Sweden\\
    \small{\texttt{\{name.surname\}@\{liu.se, kaust.edu.sa, zenseact.com\}}}  
    }

\begin{document}
\maketitle
\begin{abstract}

Computer vision systems that are deployed in safety-critical
applications need to quantify their output uncertainty.
We study regression from images to parameter values and here it is
common to detect uncertainty by predicting probability distributions.
In this context, we investigate the regression-by-classification paradigm which can represent multimodal distributions, without a prior assumption on the number of modes. 
Through experiments on a specifically designed synthetic dataset, we demonstrate that traditional loss functions lead to poor probability distribution estimates and severe overconfidence, in the absence of full ground truth distributions.
In order to alleviate these issues, we propose \emph{hinge-Wasserstein} --
a simple improvement of the Wasserstein loss  
that reduces the penalty for weak secondary modes during training. This enables prediction of complex distributions with multiple modes, and allows training on datasets where full ground truth distributions are not available.
In extensive experiments, we show that the proposed loss leads to substantially better uncertainty estimation on two challenging computer vision tasks: horizon line detection and stereo disparity estimation. The code is available at:\url{https://github.com/XZLeo/hinge-Wasserstein}
\end{abstract}    
\section{Introduction}
\label{sec:intro}


Deep neural networks have revolutionized computer vision, producing accurate predictions on a large variety of tasks.
However, for safety-critical applications, it is crucial to also quantify the uncertainty of predictions.
Observations in many tasks are inherently stochastic, \eg, low-resolution measurements or occlusions of the region of interest.
These observations are usually referred to as being subject to \textit{aleatoric uncertainty}\cite{kendall2017uncertainties}, which cannot be reduced, even given more collected data.
In many common regression tasks, \eg depth estimation and object pose estimation, the aleatoric uncertainty is usually described with multimodal distributions.
Popular training objectives for regression tasks are the $L_1$ and $L_2$ loss. 
However, these losses assume that the data follows, at least partially, a Gaussian ($L_2$) or Laplacian ($L_1$) distribution \cite{mathieu2015deep}, thereby neglecting multimodal distributions and returning the mean of multiple modes. 
\begin{figure}[t]
        \centering
        \includegraphics[width=\columnwidth]{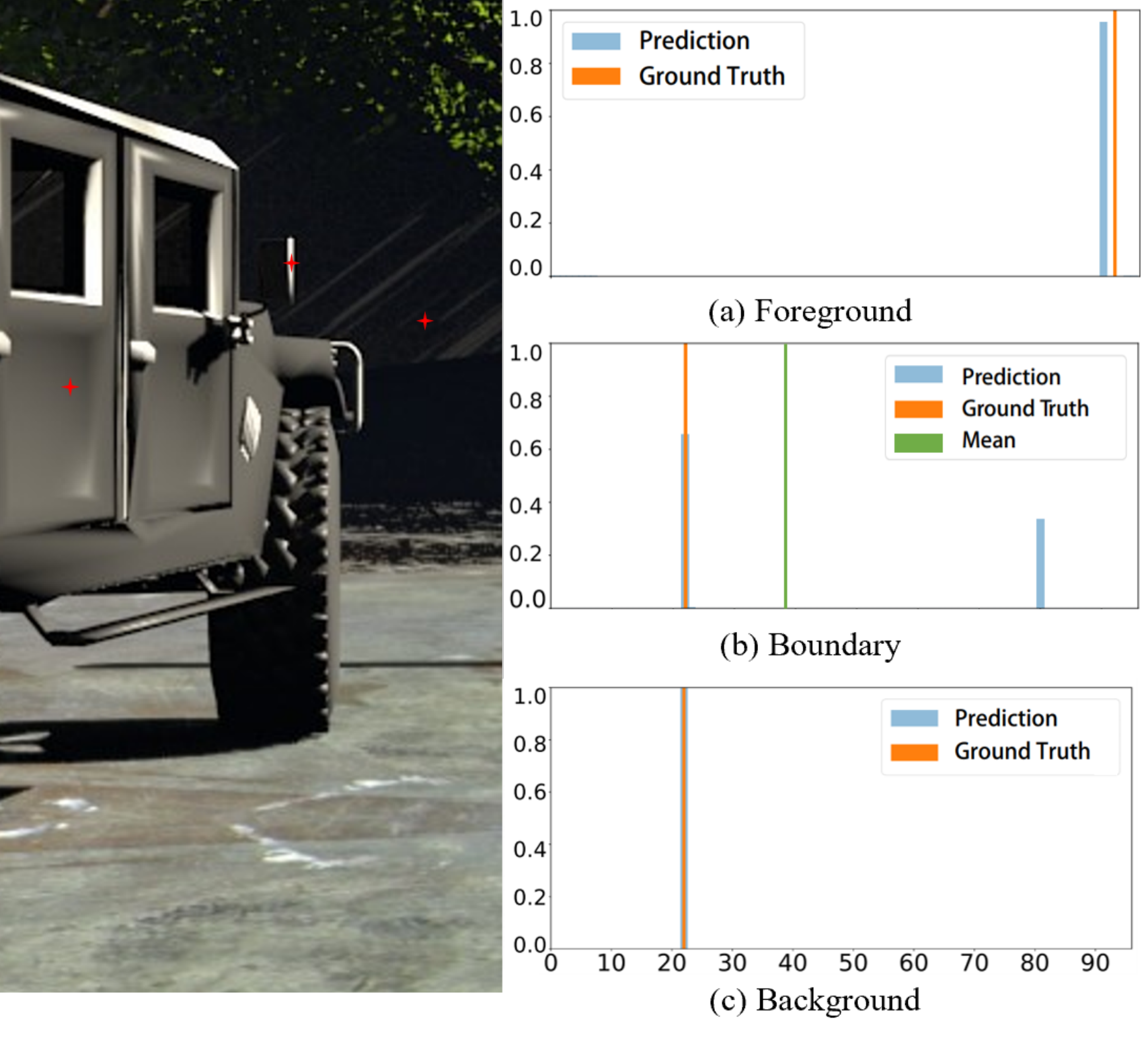}
        \vspace{-8mm}
        \caption{Edge pixels are subject to multimodal aleatoric uncertainty. Left: Three pixels are marked with red stars of the left frame. Right: Predicted disparity distributions of the three pixels, horizontal axis is disparity (same range for three subfigures), vertical axis is probability on (a) foreground (door of the car), (b) boundary pixel (edge of the mirror), and (c) background. }
 \label{fig:intro}
 \vspace{-5mm}
\end{figure}
However, in the multimodal case as shown in Fig.~\ref{fig:intro}, 
the mean can deviate from the {\it most likely mode} (which is usually the desired prediction) and is instead located in a region of low likelihood. 
Specifically, boundary pixels will have a high likelihood of both the foreground and the background, \eg, the predicted disparity map in Fig.~\ref{fig:intro} (b) shows that the disparity likelihood is 70\% at 20px, and 30\% at 80px.
The mean (38px) is by no means the real foreground disparity, and for some applications, \eg, autonomous driving, decisions based on this perception might be fatal.

A flexible approach to model multimodal distributions is \textit{regression-by-classification} (RbC) \cite{workman2016hlw,hager2021,fjg06} where the regression space is partitioned into a fixed set of bins, and the task is to predict the probability of the regressed value falling into each bin. 
\cref{fig:intro,fig:ambiguity_illustration} show examples of RbC outputs from the proposed method on stereo disparity and horizon line estimation respectively.
Furthermore, in RbC it is often argued that the Wasserstein loss should be used
\cite{Liu_2019_ICCV,hager2021,div2020wstereo}, as the regression space partitioning induces inter-class correlations. 
For instance, a predicted bin close to the ground truth mode is generally better than the one that is far away. 
Another advantage of the Wasserstein loss is that it allows for multimodal training targets, which improves regression performances \cite{div2020wstereo}.
Despite the merits of Wasserstein loss, there is a remaining issue untackled: \textbf{in most realistic tasks, such multimodal ground truth labels do not exist}.
To the best of our knowledge, none of the previous RbC works consider aleatoric uncertainty estimation in scenarios where multimodal ground truth distributions are unavailable.

To analyze the influence of different loss functions on estimating aleatoric uncertainty, we create a synthetic dataset for which the aleatoric uncertainty can be controlled.
This analysis reveals a major downside of the plain Wasserstein loss, namely poor uncertainty estimation when the full target distribution is missing.
To mitigate this, we introduce \textbf{\emph{hinge-Wasserstein}}, an improved version of the Wasserstein loss comprising a hinge-like mechanism during loss computation.
This allows weak secondary modes to exist in the predicted distribution, and is thereby able to reduce overconfidence, especially when full ground truth distributions are unavailable.
We provide an ablation study that shows the effectiveness of the proposed loss on representing multimodal distributions with unimodal targets by comparing with plain Wasserstein and sythetic multimodal targets.
We further demonstrate that the proposed loss significantly improve uncertainty estimation while maintaining the main task performances 
 on two common regression tasks, namely horizon line detection and stereo disparity estimation. 

To summarize, our main \textbf{contributions} are as follows:
\begin{itemize}
    \item We show that the plain Wasserstein loss trained with unimodal targets leads to poor uncertainty estimation. Moreover, we prove that the Wasserstein loss combined with a terminal softmax layer leads to vanishing gradients.
    \item We propose \emph{hinge-Wasserstein}, which is a easy fix to plain Wasserstein, as a training loss for regression tasks with prevalent aleatoric uncertainty. The extra computation for hinge operation is negligible.
    Additionally, we prove the proposed loss to be a proper scoring rule.
    \item Through extensive experiments, we show that hinge-Wasserstein gives better uncertainty estimation while maintaining the main task performance, especially under multimodal aleatoric uncertainty.
\end{itemize}

\begin{figure*}[t]
\centering
\begin{tabular}{@{}ccc@{}}
\includegraphics[height=3cm]{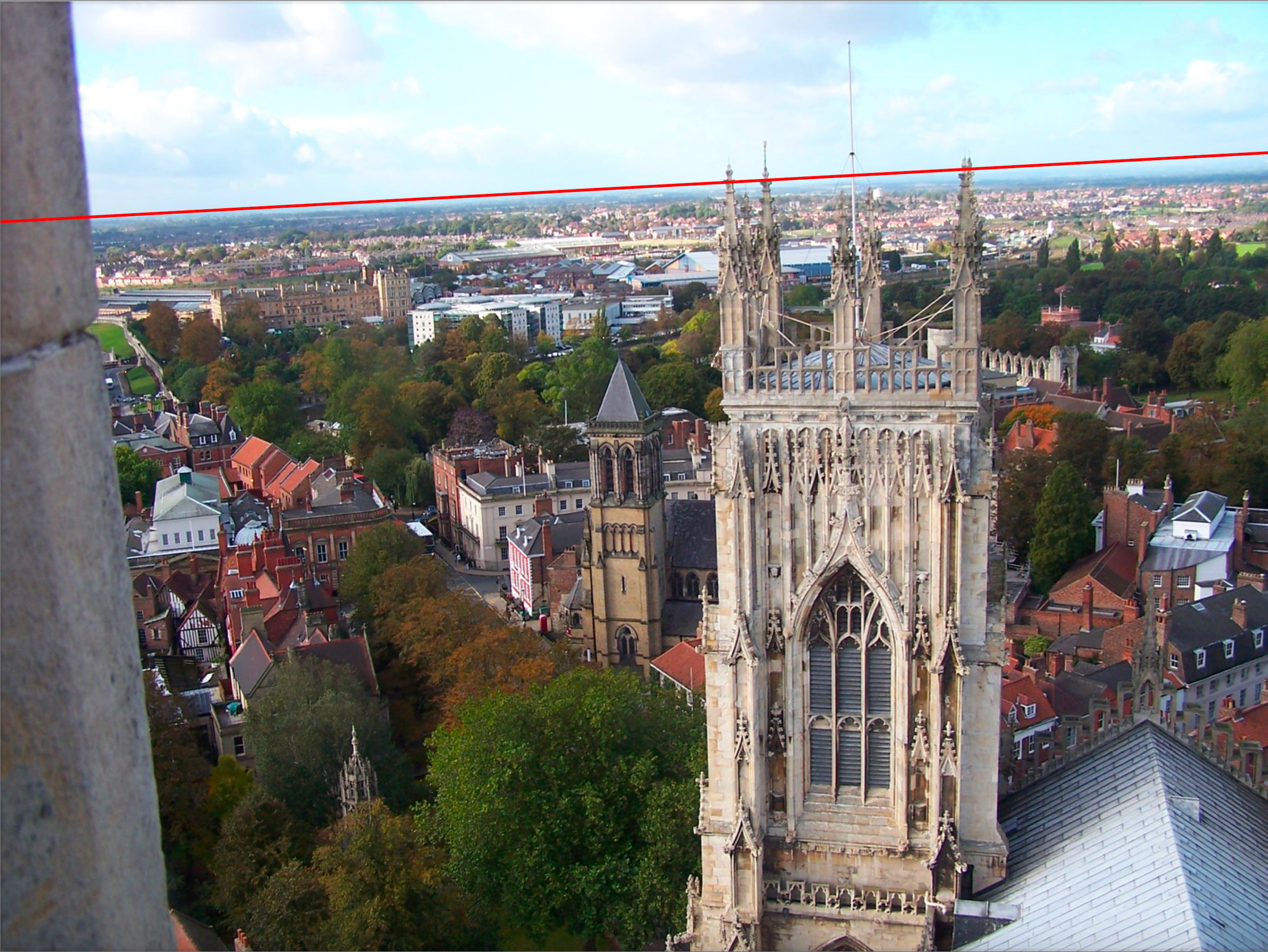} &\includegraphics[height=3cm]{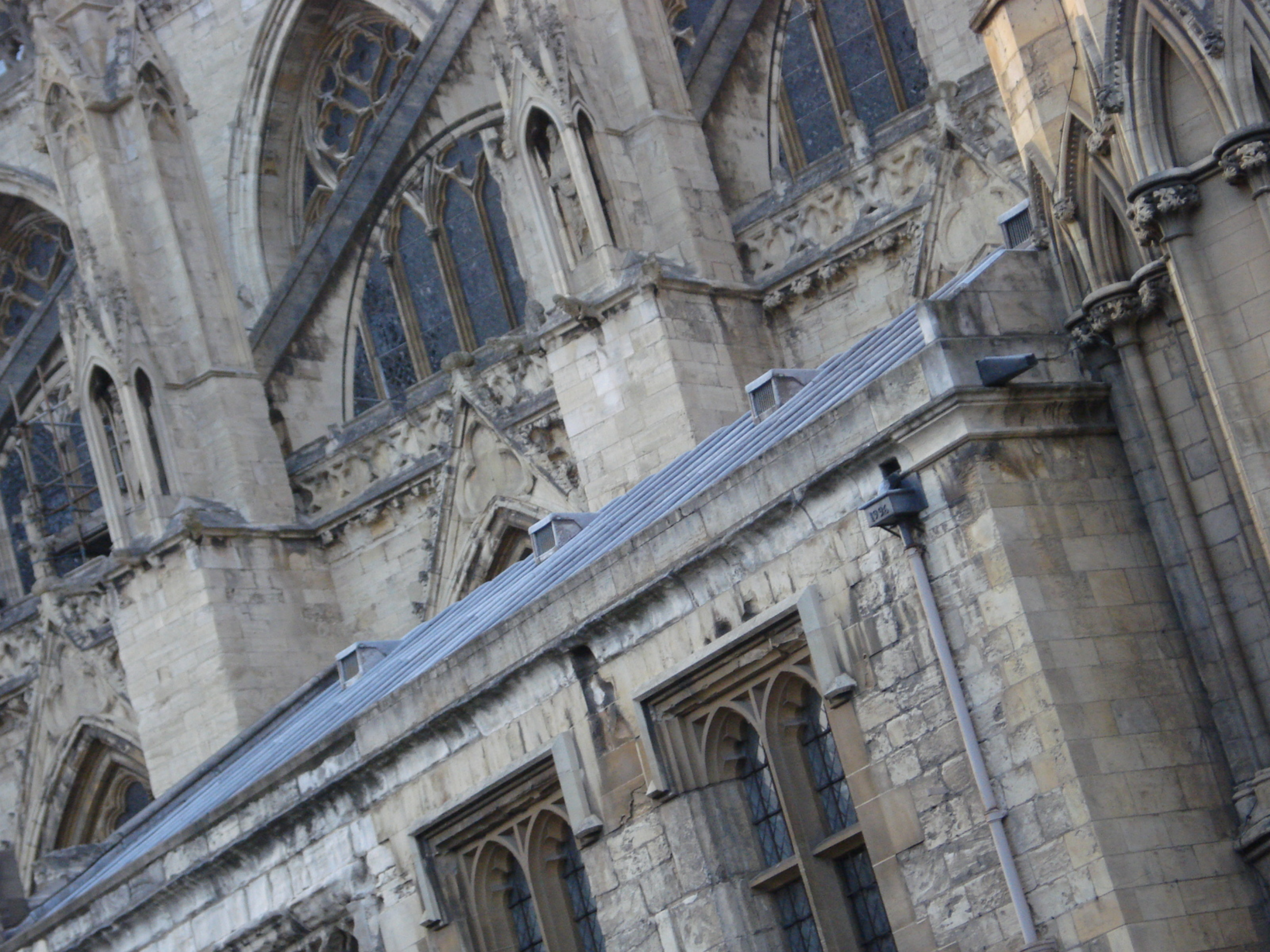} &
\includegraphics[height=3cm]{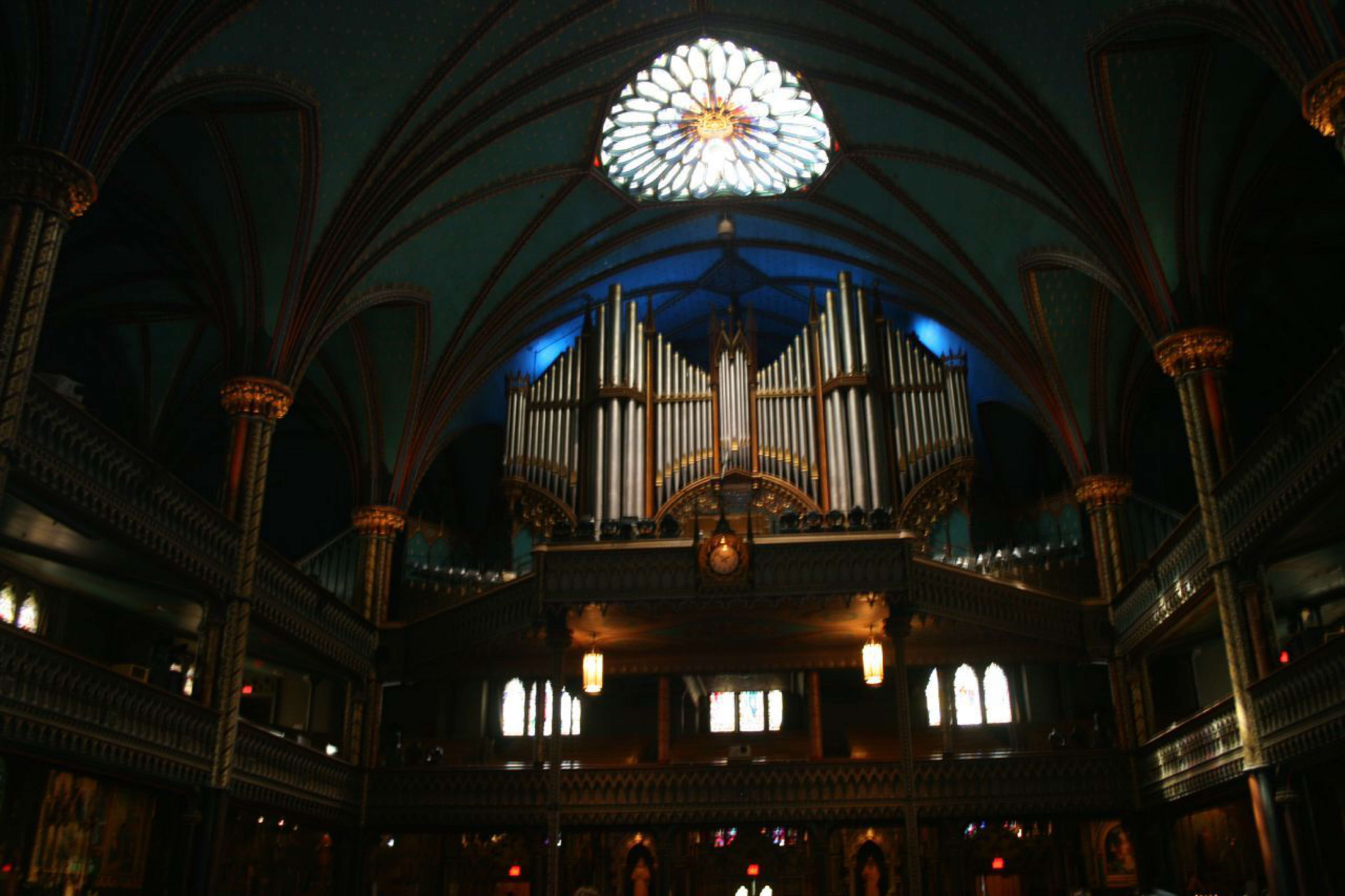} \\[-0.5ex]
\begin{tabular}{@{}lr@{}}
\includegraphics[width=1.8cm,height=1.2cm,trim = 200 190 180 210,clip]{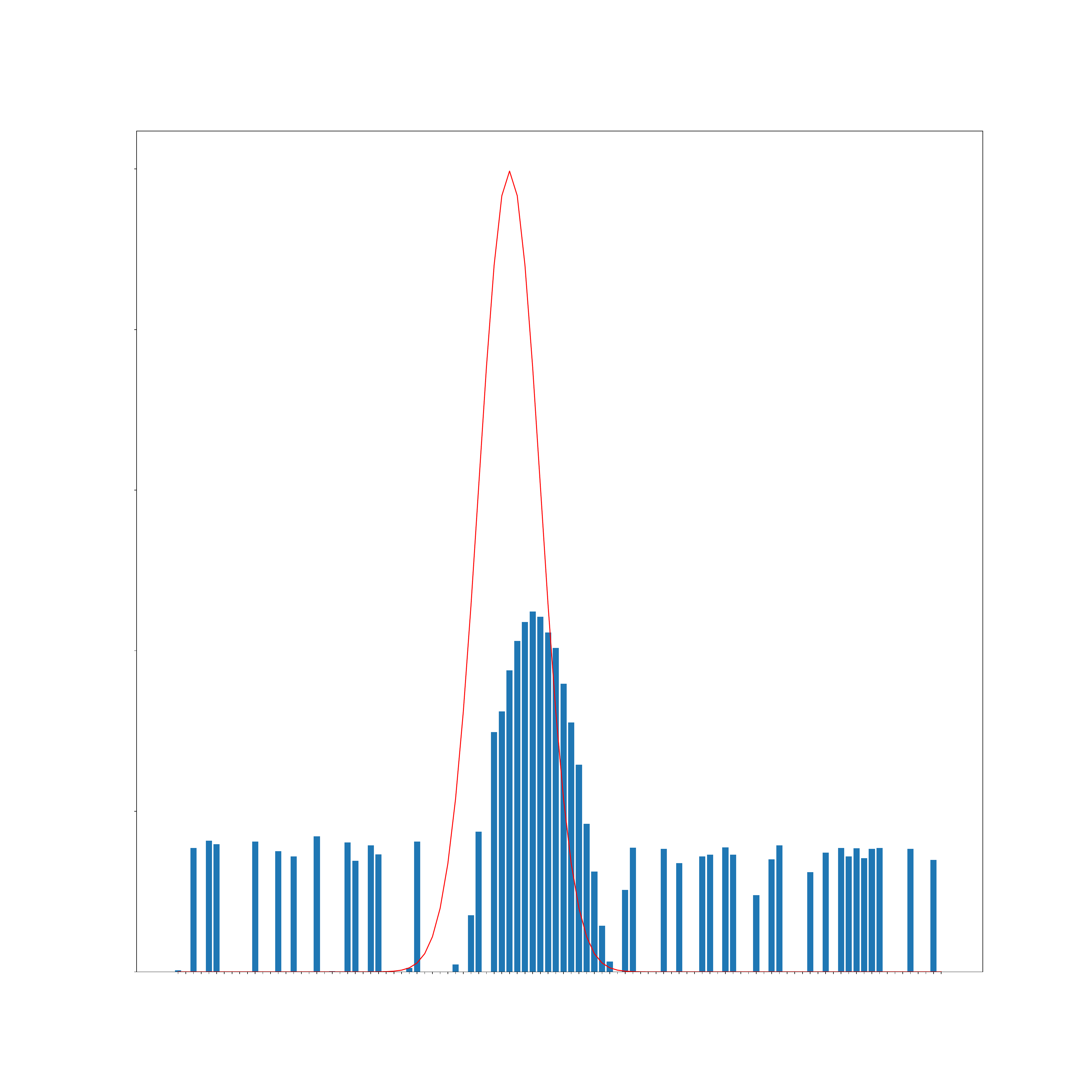} &
\includegraphics[width=1.8cm,height=1.2cm,trim = 200 190 180 210,clip]{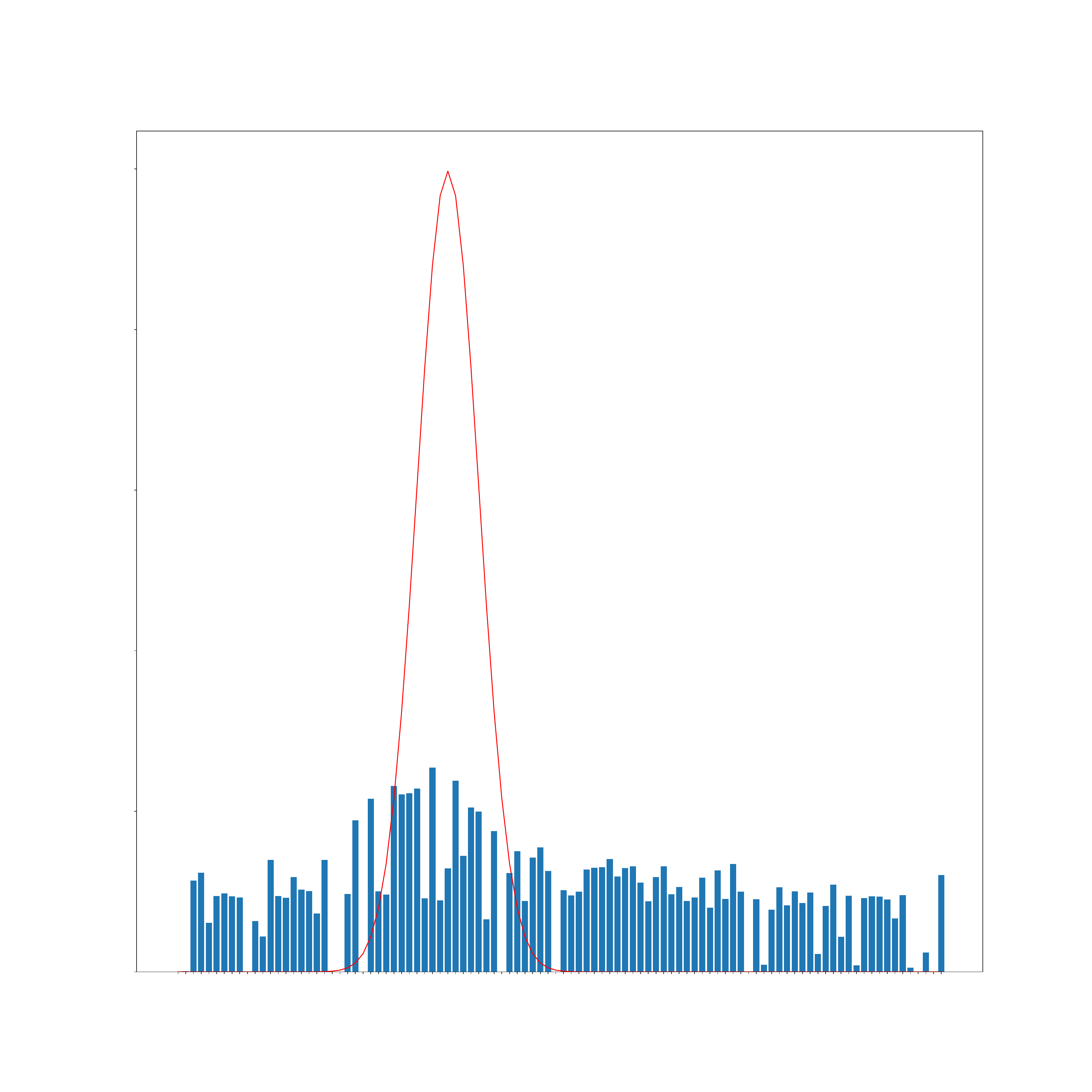}
\end{tabular} &
\begin{tabular}{@{}lr@{}}
\includegraphics[width=1.8cm,height=1.2cm,trim = 200 190 180 210,clip]{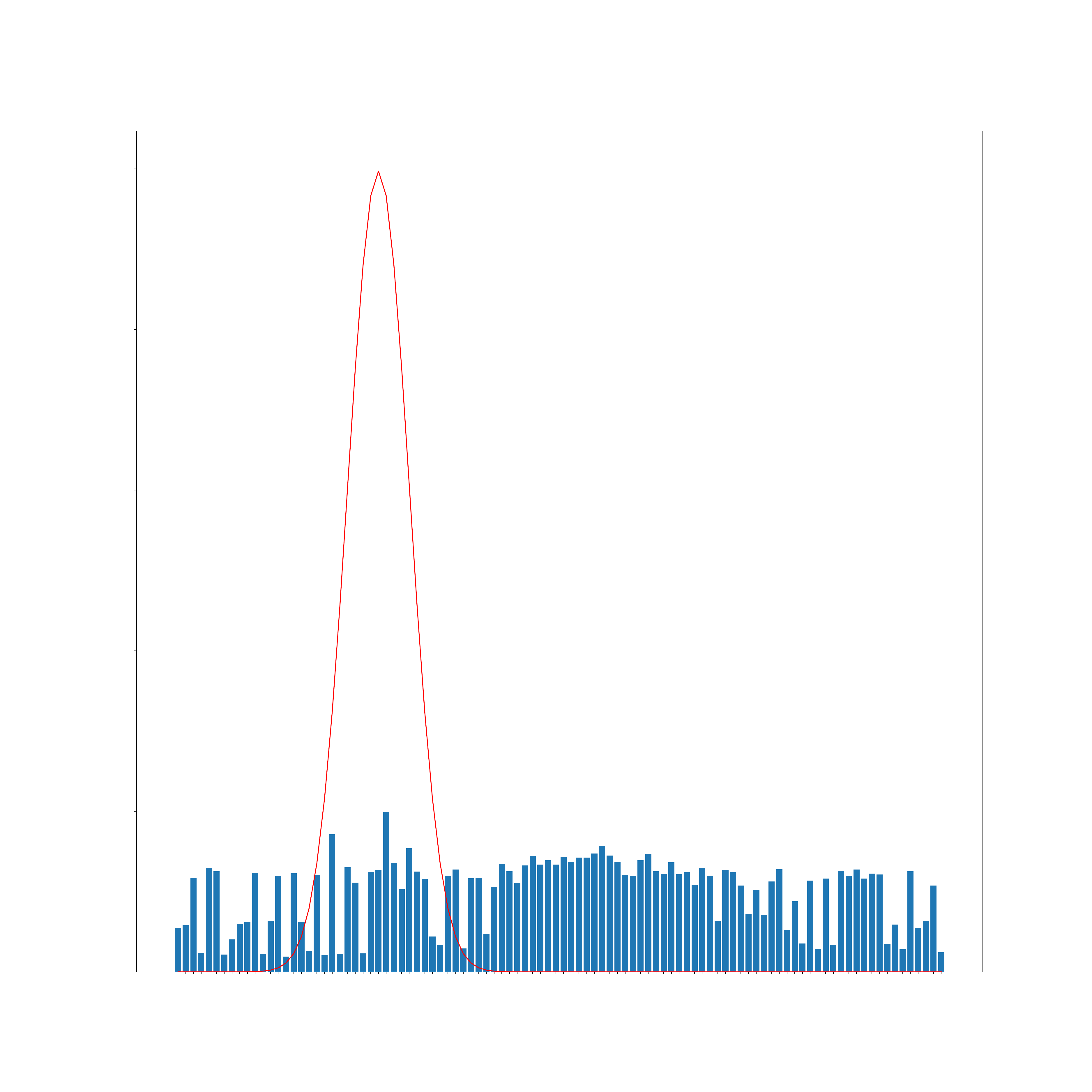} &
\includegraphics[width=1.8cm,height=1.2cm,trim = 200 190 180 210,clip]{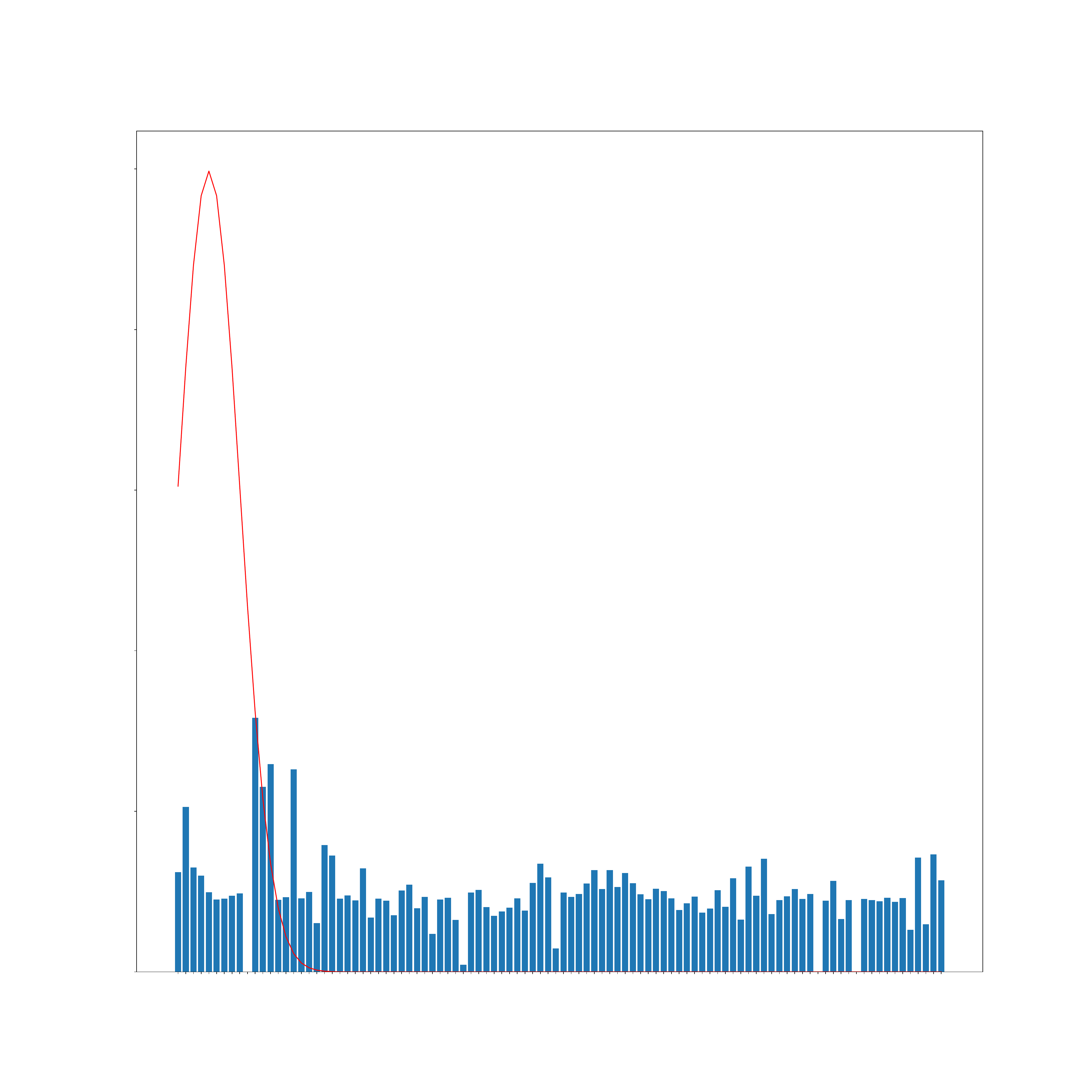} 
\end{tabular} &
\begin{tabular}{@{}lr@{}}
\includegraphics[width=1.8cm,height=1.2cm,trim = 200 190 180 210,clip]{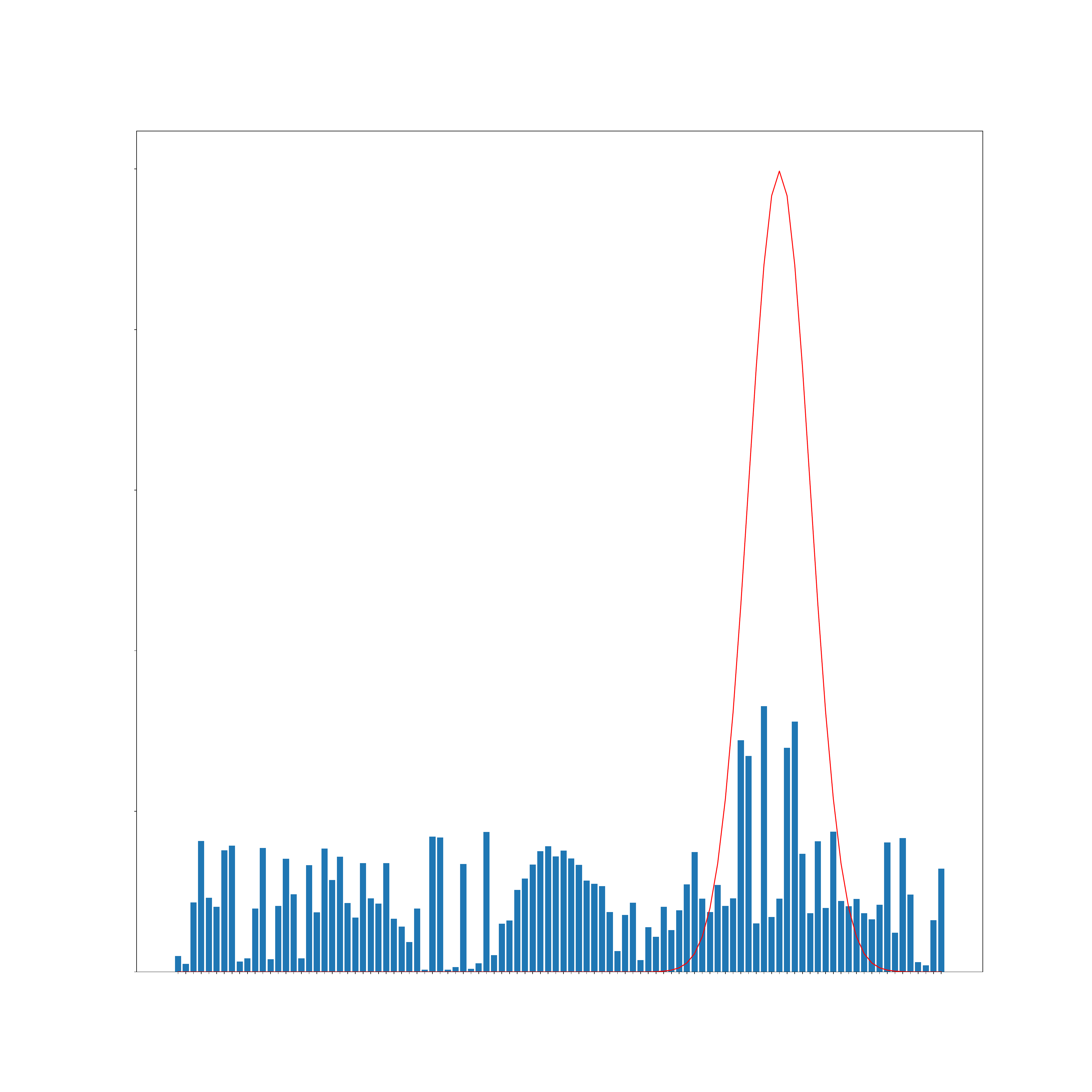} &
\includegraphics[width=1.8cm,height=1.2cm,trim = 200 190 180 210,clip]{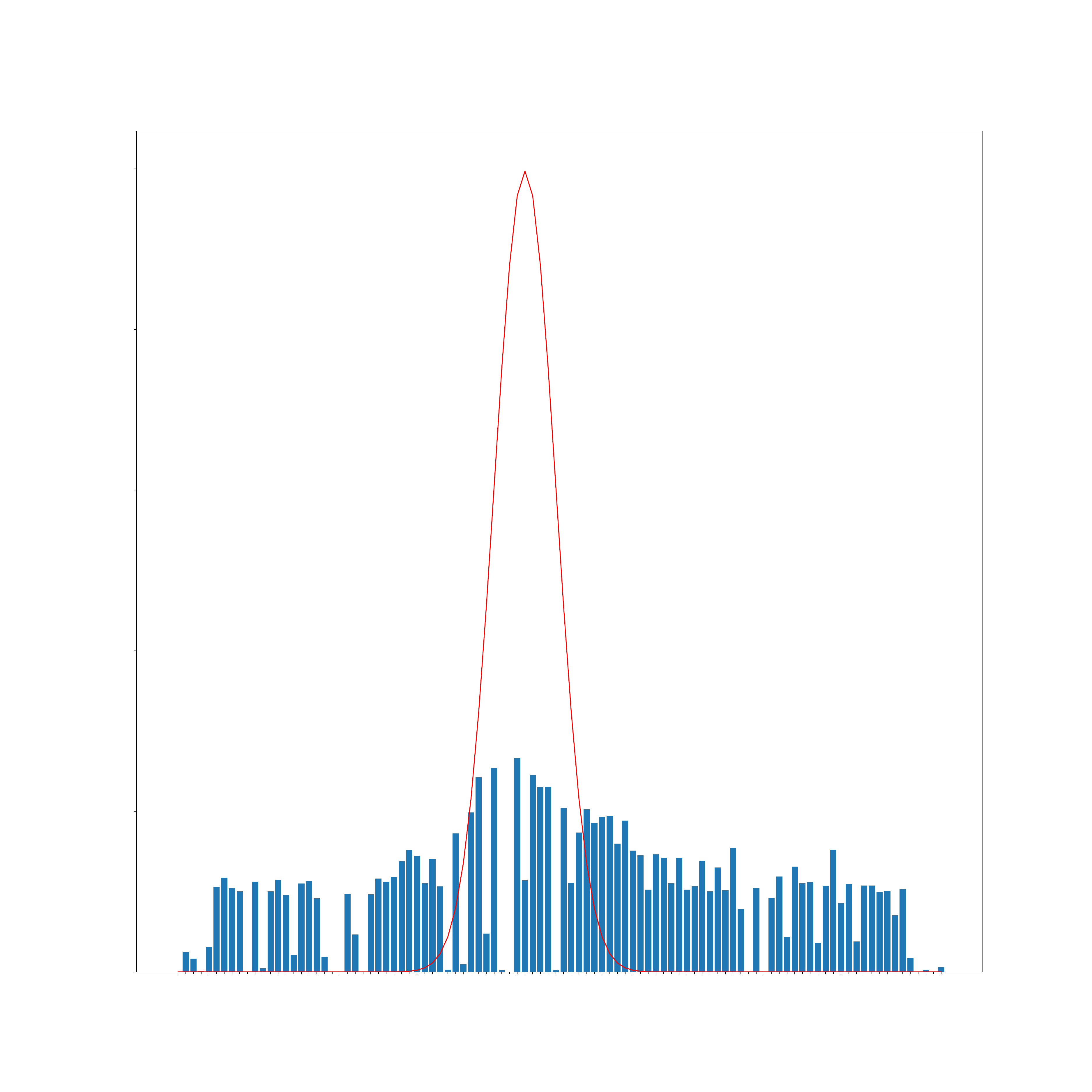}
\end{tabular}
\end{tabular}
\vspace{-4mm}
	\caption{Horizon line detection should be framed as a probabilistic regression problem due to its inherently stochastic nature. Upper Left: Image where horizon line detection is
		easy (red line) and direct regression would work. 
Upper Middle and Right: Images where
		the horizon line is ambiguous. Bottom row: Plots below
		the images show the output probability distributions for the horizon line parameters $(\alpha,\rho)$, from the proposed method. Red: Gaussian-smoothed ground truth; Blue: predicted density. Images are from the HLW dataset \cite{workman2016hlw}.}
	\label{fig:ambiguity_illustration}
    \vspace{-5mm}
\end{figure*}

\section{Related work}
\label{sec:related_work}
\parsection{Uncertainty quantification}:
Uncertainty quantification approaches can be divided into (i) {\it parametric approaches}, which are usually more task-specific; (ii) {\it ensemble methods} that are expensive but can be applied to most tasks; and (iii) {\it regression-by-classification}, which outputs full distributions \cite{hager2021}.
Parametric approaches assume an output distribution as a form of inductive bias.
The neural network is then designed and trained to predict the parameters of this assumed distribution.
The target distribution is highly specific for a certain task.
E.g., He \etal~\cite{he2019bounding} predict the parameters of a Gaussian distribution for the box width and height in object detection. A more complex distribution, Poisson multi-Bernoulli, is adopted in the work of Hess \etal~\cite{hess2022object} for probablistic object detection.
Contrarily, ensemble methods estimate uncertainty by producing a set of outputs that together characterize the uncertainty.
A well-known example is the deep ensemble~\cite{lakshminarayanan2017simple} where multiple neural networks -- trained for the same task but using different initial seeds and batch shuffling -- make predictions on each data point.
Another approach that avoids the need for multiple neural networks is Monte Carlo dropout~\cite{gal2016dropout}. Inference is conducted via multiple forward passes through a single neural network with random dropout active.
Ensemble methods typically require no modifications to the model architecture and can therefore be combined with the parametric approach to achieve better calibration~\cite{ilg2018uncertainty}.
One disadvantage with ensemble methods, however, is that they are relatively computationally expensive.

\parsection{Regression-by-classification}: The main principle for RbC is to transform regression tasks into classification tasks.
This is achieved by discretizing the continuous target variables into bins, given some prior knowledge about the problem or automatically computed ranges.
Several classical regression approaches adopted this strategy such as Support Vector Regression \cite{ZHANG2020123}, ordinal regression\cite{cheng2008neural}, and decision tree regression \cite{loh2011classification}.
In deep learning, these classical approaches were adapted to be differentiable, and thus operate in end-to-end learning frameworks~\cite{kontschieder2015deep,Papernot2018DeepKN,cheng2008neural}.
Other approaches employed this concept to solve specific problems.
Niu \etal~\cite{niu2016ordinal} tackled the problem of age estimation as ordinal regression using a series of binary classification sub-problems to utilize inherent order of labels.
Liu \etal~\cite{Liu_2019_ICCV} incorporated inter-class correlations in a Wasserstein training framework for pose estimation. 
Garg \etal~\cite{div2020wstereo} also utilized a Wasserstein training loss and proposed to have an additional offset prediction for each bin. 
They revealed the benefit of training with synthetic multimodal ground truth for improving stereo disparity. 
Despite the success of this approach, one major issue is that multimodal ground truth does not exist in real-world datasets and as we will show, training with unimodal ground truth will result in overconfidence. We therefore propose an improvement to the Wasserstein loss that is able to reduce the overconfidence, and thus allow training on real-world datasets.

\parsection{Multimodal aleatoric uncertainty}:
The term {\it aleatoric uncertainty} refers to inherent uncertainty in the observation, \eg in the form of measurement noise or limited precision, which cannot be alleviated given more data \cite{kendall2017uncertainties}. 
A task with prevalent measurement noise is horizon line detection, which is to estimate the horizon from a single image.
The images are subject to low exposure, motion blur, and occlusions, resulting in aleatoric uncertainty.
Workman \etal~\cite{workman2016hlw}
first proposed to use a CNN to directly estimate the horizon.
They also published a benchmark dataset, \textit{Horizon Lines in the Wild} (HLW), containing real-world images with labeled horizon lines.
The HLW-Net~\cite{workman2016hlw} has a GoogleNet \cite{szegedy2015going} backbone trained by soft-argmax and a cross-entropy loss, predicting line parameters.
Brachmann \etal~\cite{brachmann2019neural} achieved better performance by predicting a set of 2D points and then fits a line with RANSAC. 
Although each point has a sampling score, it fails to transfer to the probabilistic distribution of the line.
SLNet~\cite{lee2017semantic} is an architecture to
find semantic lines and surpassed the other approaches.
It proposes line pooling layers that extract line features from feature maps. 
Though a classification head of SLNet decides if a candidate line is semantic, it does not address the distribution of the line. 
In the experiment, we build upon the work of Workman \etal~\cite{workman2016hlw}, given the open-source dataset and its end-to-end model architecture.

Disparity estimation is another task with inherent aleatoric uncertainty \cite{kendall2017uncertainties}, caused by the depth discontinuities in most natural scenes. 
Especially boundary pixels between foreground and background objects are likely to be inherently multimodal, having two modes for both objects.  
Garg \etal~\cite{div2020wstereo} revealed the benefit of training with synthetic multimodal ground truth for improving stereo disparity.
However, it neither assesses the uncertainty evaluation nor addresses the unavailability of multimodal ground truth.
Häger \etal~\cite{hager2021} improve the uncertainty estimation by setting a maximum-entropy distribution as the target for those occluded pixels, and evaluate uncertainty using {\it sparsification plots}. We adopt their uncertainty evaluation and show that the proposed hinge-$W_1$ can further improve uncertainty estimation, in particular on non-synthetic datasets, where it alleviates the lack of multimodal ground truth.

\section{Method}
\label{sec:method}
We first review the theory for training Regression-by-Classification Networks in Sec.~\ref{sec:training} and the closed-form solution of the Wasserstein distance in Sec.~\ref{sec:wasserstein}. We then introduce hinge-Wasserstein loss in Sec.~\ref{sec:hinge-W} and prove it to be a proper scoring rule. Finally we intorduce uncertainty evaluation metrics in Sec.~\ref{sec:uncertainty}.

\subsection{Training regression-by-classification networks}
\label{sec:training}
In {\it regression-by-classification} (RbC), a regression variable $y\in\mathbb{R}$ is discretized into $K$ bins. 
A neural network $Z$ then predicts a conditional probability $p(y|{\bf x})$ given the evidence ${\bf x}$, which in our case is an image. 
The output of the network $\hat{p}_y = Z(x)$ in the $K$-probability-simplex is a vector where each element represents the probability that the regression variable $y$ lies in a specific interval $k$,
\begin{equation}
    \hat{p}_y[k] \approx P(v_k<y<v_{k+1}|{\bf x})\enspace,
\label{eq:bins}
\end{equation}
where $\left\{v_k\right\}_{k=1}^{K+1}$ are the bin edges.
The final regression prediction $\hat{y}$ is obtained by applying a decoding function to the output vector $\hat{p}_y$, \ie, $\hat{y}=\texttt{dec}(\hat{p}_y)$.
The decoding function $\texttt{dec}$ can be defined in different ways as explained in \cite{fjg06} depending on the exact representation of $p_y$. 
A straightforward decoding is to extract the maximum value of the output vector $\hat{p}_y$ as in \cite{workman2016hlw}.
Besides decoding, in Sec.~\ref{sec:uncertainty} we discuss computing an uncertainty measure from $\hat{p}_y$.

Training RbC networks requires defining a loss on the output vector $\hat{p}_y$ with respect to some ground truth annotation $p^\ast_y$. Usually only a single continuous value, $y^\ast$, is given as ground truth. This can be interpreted as the corresponding distribution being a Dirac impulse, $p_y^\ast = \delta[y-y^\ast]$. A natural choice of loss for this setting is the NLL of $y^\ast$ under the predicted distribution, $\hat{p}_y$~\cite{workman2016hlw}. Kendall \etal~\cite{kendall2017stereo} instead propose to minimize the decoding error, $\hat{y}$. 
But Häger \etal~\cite{hager2021} argue that this can lead to a biased output when aleatoric uncertainty is present.

\subsection{Wasserstein loss}
\label{sec:wasserstein}

The output labels in a RbC network have an inherent order, see \eqref{eq:bins}. The errors caused by predicting a bin close to the correct one are thus less severe than predicting one farther away. This detail is ignored by the standard cross-entropy loss, KL divergence,.etc, and it has thus been argued that the Wasserstein loss is a better fit for regression-by-classification \cite{Liu_2019_ICCV}.
The Wasserstein loss between a predicted distribution $p(y)$ and the
ground truth $q(y)$ over some variable $y$ is defined as
\begin{equation}
	W_m(p,q)=\inf_{\gamma\in\Gamma(p,q)} \int\int |y'-y|^m\gamma(y,y')dydy'\enspace,
\label{eq:wasserstein_full}
\end{equation}  
where $\Gamma(p,q)$ is the set of  all possible
transport plans that take $q$ to $p$. 
For $m=1$ the search over transport plans in \eqref{eq:wasserstein_full} can be avoided
\cite{thorpe18,thorarinsdottir13}, and in \eqref{eq:bins}, the output distribution is discretized, which leads to:
\begin{equation}
    W_1(p,q)=\sum_{k=1}^K |P[k]-Q[k]|\enspace.
	\label{eq:wasserstein_closed}
\end{equation}
Here, $P$ and $Q$ are the CDFs of $p$ and $q$
respectively, \ie $P[k]=\sum_{k=1}^K p[k]$ and 
$Q[k]=\sum_{k=1}^K q[k]$.
Further, for non-synthetic datasets e.g.\ \cite{workman2016hlw}, the annotations are Dirac distributions, $q(y)=\delta(y-y^\ast)$. This also avoids the search over transport plans \cite{garg2020}:
\begin{equation}
	W_m(p_y,q)=\left(\sum_k p_y[k](\hat{y}(p)-y^\ast(q))^m\right)^{1/m}.
	\label{eq:wasserstein_dirac}
\end{equation}
\parsection{Softplus Activation}:
 Ensuring that $\hat{p}_y$ in \eqref{eq:bins} is on the $K$-probability-simplex is most commonly achieved via the softmax function. 
 Below, we show that vanishing gradients occur for the combination of a softmax layer and the $W_1$ distance with Dirac ground truth. Softmax, commonly used as the final layer for multi-class classifiers, is defined in \eqref{eq:softmax}. 
\begin{equation}
    g_i(\textbf{z}) = \frac{e^{z_{i}}}{\sum_{j=1}^N e^{z_{j}}}\,, \ \ \textup{for}\ \ i=1,2,\dots,N    ,
\label{eq:softmax}
\end{equation}
where $N$ is the number of classes, and $\textbf{z}=[z_i,i=1,2,\dots,N]$.
Softmax normalizes 
its output to sum to one. Thus, we can treat the output 
$\mathbf{g}(\mathbf{z})=[g_1(\mathbf{z}),g_2(\mathbf{z}),\ldots,g_N(\mathbf{z})]$
as the predicted categorical distribution for each class. The ground truth label is $\delta[j-j^\ast]$,
where $j^\ast$ is the true class index. In this case, the $W_1$ distance can be simplified as
\begin{equation}
    W_1(\delta[j-j^\ast], \textbf{g}(\textbf{z}))=\sum_{i} \lvert i-j^* \rvert g_i(\textbf{z}) ,
    \label{eq:w1_dirac}
\end{equation}
 The partial derivative of \eqref{eq:w1_dirac} with respect to $z_k$ is 
\begin{equation}
    \label{eq:full}
     \frac{\partial  W_1(\delta[j-j^\ast], \textbf{g}(\textbf{z}))}{\partial z_k} = g_k(\textbf{z})(\lvert k-j^* \rvert-\sum_{i}\lvert i-j^* \rvert g_i(\textbf{z}))
    .
\end{equation}
There are two problematic cases for \eqref{eq:full}:

Case 1: Low initial value for the correct bin. For the correct bin, where $k=j^\ast$, $\lvert k-j^* \rvert$ in \eqref{eq:full} is zero. Further, if $g_k({\bf z})\approx 0$, we have a small, but always negative, contribution from the sum, since $g_k({\bf z})$ is a factor of \eqref{eq:full}.  Thus, if $g_k({\bf z})$ starts out as a small value, it will be hard to change the value. In the limit case, where $g_k({\bf z})=0$ the gradient is zero, resulting in no learning at all.

Case 2: Dominant, but incorrect mode. If there is a dominant, but incorrect mode $m\not = j^\ast$, the contribution to the gradient will also be low. In the limit we have  $g_m(\textbf{z})=1$, $g_i(\textbf{z})= 0$, for $i\not = m$ (as $\sum_i{g_i({\bf z})=1}$). In this case, both terms in (\ref{eq:full}) also become close to 0, for all values of $k$. 
Thus, whenever there is a peak of a wrong class close to 1, this will also cause vanishing gradients.\\
We instead follow Häger \etal~\cite{hager2021}, applying softplus function followed by an $l_1$ normalization, which the authors found beneficial for regression-by-classification.
 Our experiments in Sec.~\ref{sec:disparity} show that simply switching to softplus will slightly improve uncertainty estimation and regression. 


\subsection{Hinge-Wasserstein loss}
\label{sec:hinge-W}
Most of the datasets used in computer vision are annotated with unimodal ground truth, \ie, instead of full conditional density annotations,
only the most likely output is provided as the training target. 
When we use a loss that rewards output of only a single mode at the annotation, we effectively discourage the output to represent
aleatoric uncertainty, which in turn causes overconfidence. In the extreme case, when there is no evidence to support {\it any} hypothesis, the output in \eqref{eq:bins} should be uniformly distributed.
~\footnote{Note that what we expect to see is the dataset prior. The bins, however, are usually chosen as quantiles of the training set, which results in a uniform distribution over the bins.}

An intuitive way to use unimodal ground truth to learn probability density outputs is to discount the loss for all bins by a margin
$\gamma_W$, and not penalize bins that are below this level. This will allow aleatoric uncertainty in the input to be represented
in the output. This is similar to the hinge losses from support vector machines and to the triplet 
loss used in contrastive learning \cite{schroff2015}, where it only matters if the distance to a negative sample is larger up to a certain point.
Similarly, we allow incorrect bins to be non-zero, as long as they are sufficiently below the level of the main mode.
In detail, the predicted probability density will be reduced by the threshold, passed through a ReLU and then re-normalized to sum to one,
\begin{align}
	\tilde{p}[k] &= \max(p[k] - \gamma_W, 0)\enspace,\\
	\overline{p}[k] &= \frac{\tilde{p}[k]}{\sum_{k}\tilde{p}[k]}\enspace.
\end{align}
The loss is then defined as the Wasserstein distance between the the renormalized
probability densities and ground truths,
\begin{equation}
  \mathcal{L}(p, q) = W_1(\overline{p}, q)\enspace.
\label{eq:loss_hinge_emd}
\end{equation}
We call this new loss the \textit{hinge-Wasserstein} loss (abbreviated hinge-$W_1$).
Note that (\ref{eq:loss_hinge_emd}) works for both implementations of $W_1$ as described by (\ref{eq:wasserstein_closed}) and (\ref{eq:wasserstein_dirac}), i.e., it allows both a Dirac training target and a full distribution target.
The parameter $\gamma_W$ depends on the total number of output bins, and is normally set as $\gamma_W=1/K$.
This can be interpreted as the hinge-Wasserstein loss allowing a random guess (which would correspond to a uniform distribution).

\parsection{Proper scoring rule}
A {\it scoring rule} $S(p_\theta,q)$ \cite{gneiting07} is a function that evaluates
the quality of a predictive distribution $p_\theta(y|{\bf x})$, with
respect to a true distribution $q$. 
A {\it proper scoring rule} \cite{gneiting07} should satisfy:
$
S(p_{\theta}, q)$$ \leq$ $S(q, q)$$,\forall p_\theta\enspace,
\label{eq:proper_scoring_rule}
$
with equality if $p_\theta$$=$$q$. If the inequality is strict for all $p_\theta$$\not$$=q$, $S$ is \textit{strictly proper.}
It has been shown that $W_m$ is not proper for finite samples \cite{thorarinsdottir13}.
However, we use discrete distributions with $m=1$, which can be proved to be proper.

\noindent{\bf Proposition:} {\it Discrete $-$$W_1$ is strictly proper.}

\noindent{\bf Proof by contradiction}: 
First we note that $-W_1(q,q)=0$ for any PDF $q$, see \eqref{eq:wasserstein_closed}.
Thus, setting $S=-W_1$ results in the requirement:
$W_1(p_\theta,q)\geq 0$ with equality iff $p_\theta=q$. 
\noindent\textbf{Assumption}: There exists a discrete distribution $p$ that is different from
$q$, for which $W_1(p, q)=0$.
In $p$, at least two bin values must be
different from those in $q$, as $\sum_k p[k]=1$ must hold. 
Denote their bin indices by $m_1$, $m_2$, where $m_1<m_2$, without
loss of generality.

\noindent{\bf Contradiction:} Consider \eqref{eq:wasserstein_closed} is a sum of non-negative values (since the absolute value). According to the assumption we have $p[m_1]$$\not=$$q[m_1]$,
and thus, $D[m_1]>0$ and as a consequence $W_1(p,q)$$>$$0$.
This is contradictory to the assumption. For cases where more than two
bin values are different, we will get more non-zero terms $D[k]$
in $S$ contributing to the contradiction. This
proves the equality uniqueness. \hfill $\square$

Based on the proposition above, it can be easily shown that \textit{hinge-Wasserstein is a proper scoring rule but not strictly proper} because with the hinge mechanism $p_\theta$ that achieves the optimal $w_1$ distance is not unique.

\subsection{Predictive uncertainty evaluation}
\label{sec:uncertainty}
We evaluate uncertainty estimation capabilities using \emph{sparsification plots} and \emph{CRPS}.

\parsection{Sparsification Plots}: 
A \textit{sparsification plot} \cite{aodha_tpami13,ilg2018uncertainty} assesses how well the predicted uncertainty coincides with the output error.
This requires a 
scalar uncertainty value $u$, which is a function of the network probability density output,
$\hat{p}_y=Z({\bf x})$,
\begin{equation}
	u=f(\hat{p}_y)\enspace.
	\label{eq:uncertainty_measure}
\end{equation}
A {\it sparsification curve} is a plot of the mean absolute error (MAE) as a function of a fraction $p$ of samples that have been removed. The removed samples are those with the highest predicted uncertainty according to $u$ in \eqref{eq:uncertainty_measure}.
Similarly, an {\it oracle curve} is created by instead removing the fraction $p$ of samples with the highest absolute errors.
The oracle curve as a monotonically decreasing function indicates the lower bound of sparsification curves.
Both curves are normalized by the MAE on all the test set samples, i.e.\ they start at $(0,1)$.
See Fig.~3 in \cite{ilg2018uncertainty} for an example of sparsfication plots.
By plotting the vertical distance between the two curves, defined as the sparsification error, against $p$, we get the sparsification error plot as in Fig.~\ref{fig:sparsification}.
We use {\it area under sparsification error} (AUSE) to quantitatively evaluate uncertainty estimation to fairly compare different approaches with different oracles.
A common choice for $u$, is the Shannon entropy \cite{hager2021} \cite{namdari2019review} \cite{rudnicki2011shannon} of $\hat{p}_y$,
\begin{equation}
  u_H(\hat{p}_y)=H(\hat{p}_y)=-\sum_{k=1}^K \hat{p}_y[k]\log{\hat{p}}_y[k]\enspace.
  \label{eq:entropy}
  \vspace{-1mm}
\end{equation}
We show in the supplement the comparison between entropy and other uncertainty measures, e.g., variance, maximum value of the predicted histogram, and \eqref{eq:entropy} returns best AUSE.
Hence, we show experiment results with entropy.

\parsection{CRPS}: While AUSE measures the correlation between uncertainty magnitude and prediction error, it disregards the shape of each predicted probability distribution. As the goal is to predict multimodal distributions, we can use a \textit{proper scoring rule} to assess this.
A proper scoring rule addresses calibration and sharpness simultaneously \cite{gneiting2007probabilistic}. 
Here we extend \textit{continuous ranked probability score} (CRPS) to allow multimodal ground truth. Specifically, the step function in CRPS is replaced with the cumulative distribution of equally-weighted Dirac mixtures:
$
    q(y) = \frac{1}{N}\sum_i^N\delta(y_i^\star)\enspace,
$
\begin{equation}
    \textup{CRPS}(p_{\theta}, q) = \int_{-\infty}^{\infty}\left|p_\theta(y)-q(y)\right|^2dy\enspace.
\end{equation}
Note that this metric only works when multimodal ground truth distribution is available. 



\section{Experiments}
\label{sec:experiments}

We analyze the performance of the proposed hinge-$W_1$ loss under multimodal aleatoric uncertainty. 
First, we construct a synthetic dataset, where we control the presence of aleatoric uncertainty, to show that using the negative log-likelihood or Wasserstein losses lead to overconfidence.
Our proposed hinge-$W_1$, in contrast, reports uncertainty whenever uncertainty is present. Next, we evaluate our hinge-$W_1$ loss on two separate regression tasks: horizon line regression on \textit{Horizon Lines in the Wild} (HLW) (Sec.~\ref{sec:hlw_evaluation}), and stereo disparity estimation on \textit{Scene Flow Datasets} (Sec.~\ref{sec:disparity}). 


\subsection{Synthetic dataset}
\label{sec:synthetic_dataset}
\begin{figure}[t]
        \centering
	\begin{tabular}{@{}c@{\quad}c@{\quad}c@{}}
        \includegraphics[width=4.2cm]{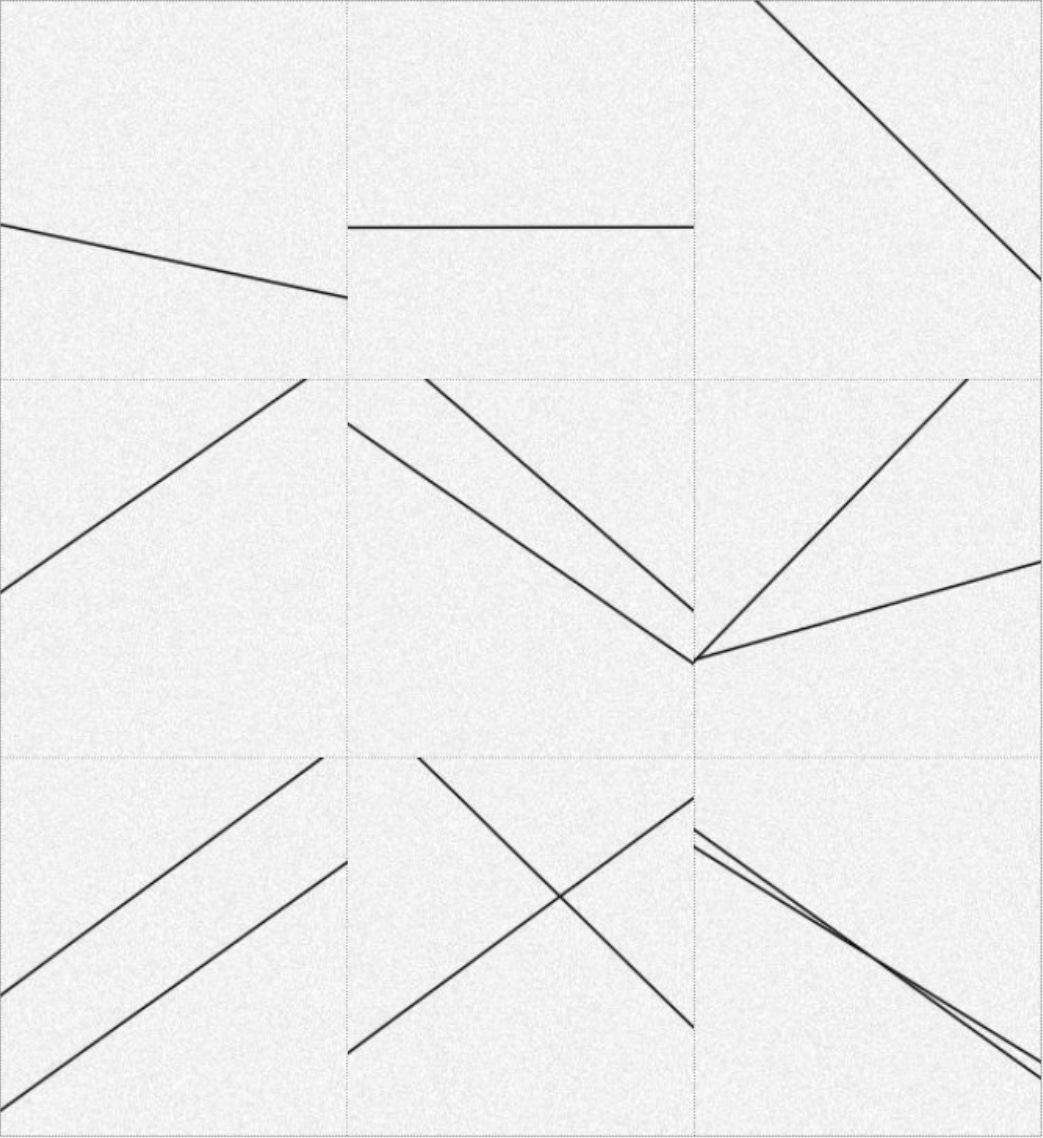} &
        \includegraphics[width=1.5cm]{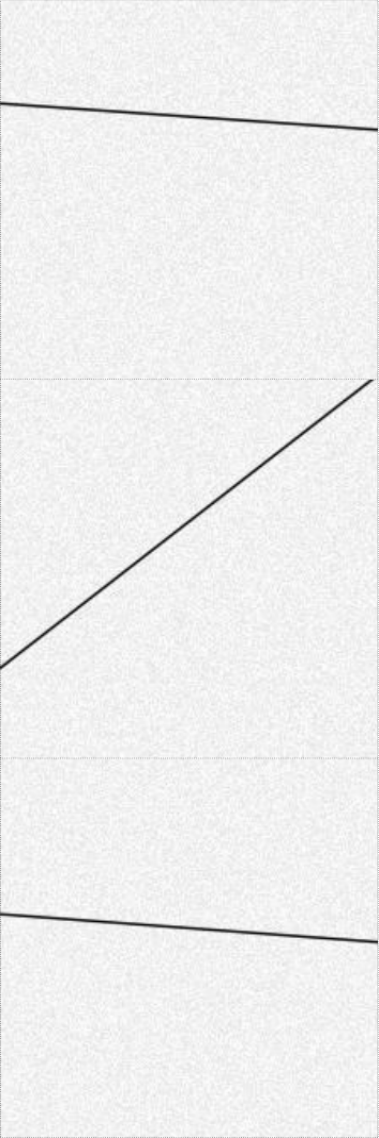} &
        \includegraphics[width=1.5cm]{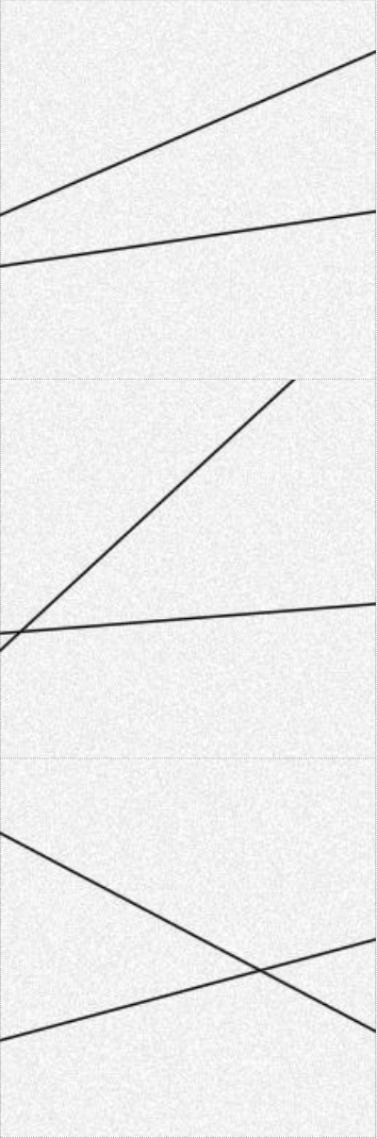} \\
        (a) Training set & (b) clear  & (c) ambig. 
        \end{tabular}
        \vspace{-3mm}
	\caption{Example images from the synthetic dataset with controllable aleatoric uncertainty: (a) training set, one or two lines per image, one line in the annotation for unimodal training; both lines in the annotation for multimodal training; (b) test set 1, one line per image; and (c) test set 2, two lines per image.}
 \label{fig:synthetic}  
 \vspace{-6mm}
\end{figure}

\parsection{Dataset}: 
To create a controlled environment where we can analyze the behaviour of our proposed approach, we choose a simplistic yet representative regression problem -- parameter estimation for lines in noisy images.
Aleatoric uncertainty is introduced in the form of multiple lines in the images as shown in Fig.~\ref{fig:synthetic}.
One or more randomly generated lines are rendered on a background with Gaussian noise.
These image-line pairs can be used for supervised training.
We can easily control the aleatoric uncertainty amount by varying the fraction of images with multiple lines. 

The training set contains 2000 clear images with one line and 2000 ambiguous images with two lines. 
We create two test sets: The clear set shown in Fig.~\ref{fig:synthetic} (b), contains 500 images with one line per image;
the ambiguous set shown in Fig.~\ref{fig:synthetic} (c), contains 500 images with two lines per image. 
To imitate real-world tasks, a randomly-picked line in the ambiguous images is labeled as the unimodal ground truth.
By labeling both lines in the ambiguous images, we also create multimodal ground truth distributions for $\alpha$ and $\rho$ (called MM GND in the experiments).

\parsection{Implementation details}: A line is often parameterized as a slope and an offset $(\alpha, \rho)$, which define the set of pixel coordinates $(x,y)$ as
$
	\begin{pmatrix} -\sin\alpha & \cos\alpha & -\rho\end{pmatrix}^T\begin{pmatrix}x & y& 1\end{pmatrix} = 0\,, \ (x,y)\in\Omega\enspace,
$
where $\Omega$ is the set of image coordinates.
To make the representation 
unique, we also restrict the parameters to $\alpha \in [-\pi/2, \pi/2]$ and $\rho \in [0, \infty)$.
Then $\alpha$ and $\rho$ are discretized into $K$$=$$100$ bins.
The first 99 bins are linearly-spaced from the minimum value of training samples to the maximum, and the last bin is from the maximum to infinity as in \cite{workman2016hlw}.
All models use a \textit{Resnet18}~\cite{he2016residual} as the backbone network, with the last layer replaced by a linear layer followed by two prediction heads, each of which consists of a softplus and normalization as in Sec.~\ref{sec:wasserstein}, for the slope and the offset respectively.
We assume that $\hat{p}_\alpha$ and $\hat{p}_\rho$ are independent. 
We conduct an ablation study on NLL, $W_1$ and hinge-$W_1$ losses with both (i) unimodal and (ii) multimodal ground truths.
For computation, we follow \cite{hager2021} and apply small Gaussian smoothing around the unimodal target.
Note that due to the use of an argmax decoding (inherited from \cite{workman2016hlw}) there always exists a quantization error given by the bin sizes, and thus an upper bound on the AUC. This error could be reduced by a more advanced decoding (see \cite{fjg06}) or by adding an offset prediction branch (see \cite{garg2020}). 

\parsection{Metrics}:
To evaluate line regression performance, we employ a metric that is commonly used in the similar task of horizon line detection, the {\it horizon detection error}, proposed by Barinova \etal \cite{barinova2010geometric}. 
It is calculated as the maximum vertical distance between ground truth and predicted lines in the image, normalized by the height of the image.
The cumulative histogram of the horizon detection error is often used to assess the error distribution for the test set, and the area under the curve (AUC) is commonly reported as a summary.
For uncertainty evaluation, we apply AUSE and CRPS (Sec.~\ref{sec:uncertainty}) separately on the line parameters.

\parsection{Results}: 
Fig.~\ref{fig:density_aleatoric} shows qualitative examples of inferences. 
The hinge-$W_1$ captures the multimodality in the input in most cases, except for Fig.~\ref{fig:density_aleatoric} (b), the {\it metameric case} \cite{fjg06}, where the modes are too close and interfere.
Table.~\ref{tab:synthetic} shows the quantitative results of both line regression performance and uncertainty evaluation on the two-line test set.
The model trained with plain Wasserstein and unimodal targets shows overconfidence with high AUSE and CRPS, which indicates plain $W_1$ severely omits secondary modes.
Though usually unavailable, adding multimodal targets to training largely improves the uncertainty scale (AUSE) and predicted distribution (CRPS) while slightly improves regression. 
While increasing hinge value $\gamma_W$, AUSE and CRPS keep improving until the $\gamma_W$ is too large. The AUSE of hinge Wasserstein with $\gamma_W$$=$$0.015$ are close to those with multimodal targets.
A large hinge damages the training because there are few bins in the predicted density above the hinge at the beginning, thus no gradient to backpropagate.

\subsection{Horizon lines in the wild}
\label{sec:hlw_evaluation}
\parsection{Dataset}:
To show that the findings in Sec.~\ref{sec:synthetic_dataset} for the synthetic dataset transfer to a related task in the real world, we evaluate our approach on the challenging \textit{Horizon Lines in the Wild} (HLW)~\cite{workman2016hlw} benchmark.
HLW is a large dataset of real-world images
captured in a diverse set of environments, with horizon lines annotated using {\it structure from motion}.
It contains 100553 training images and 2018 test images. 

\parsection{Implementation Details}:  Labels for the horizon line slope and offset are continuous in HLW.
They need to be discretized into $K$ bins respectively and use the one hot vector of bin index as a ground truth $q_y$.
The $K=100$ bins are chosen to be approximately equally likely to occur, by linearly interpolating the cumulative distribution function of the corresponding parameter over the training set.
We follow the training procedure in \cite{workman2016hlw} and replace the backbone with Resnet18. 
We apply hinge-$W_1$ as the training loss and compare with the original NLL loss, an ensemble of $5$ instances (as in \cite{lakshminarayanan2017simple}) trained with NLL loss and Plain $W_1$.

\parsection{Metrics}: We use the same regression metric (AUC) and uncertainty evaluation metric (AUSE) as in Sec.~\ref{sec:synthetic_dataset} and Sec.~\ref{sec:uncertainty}.
The only exception is that CRPS is not computed because multimodal ground truth targets are unavailable. 

\begin{figure}[t]
\centering
\resizebox{0.99\linewidth}{!}{
\begin{tabular}{@{}c@{}c@{}}
\sffamily{ $\alpha$ entropy} & \sffamily{ $\rho$ entropy} \\
\includegraphics[height=4.5cm,trim = 10 0 30 35,clip]{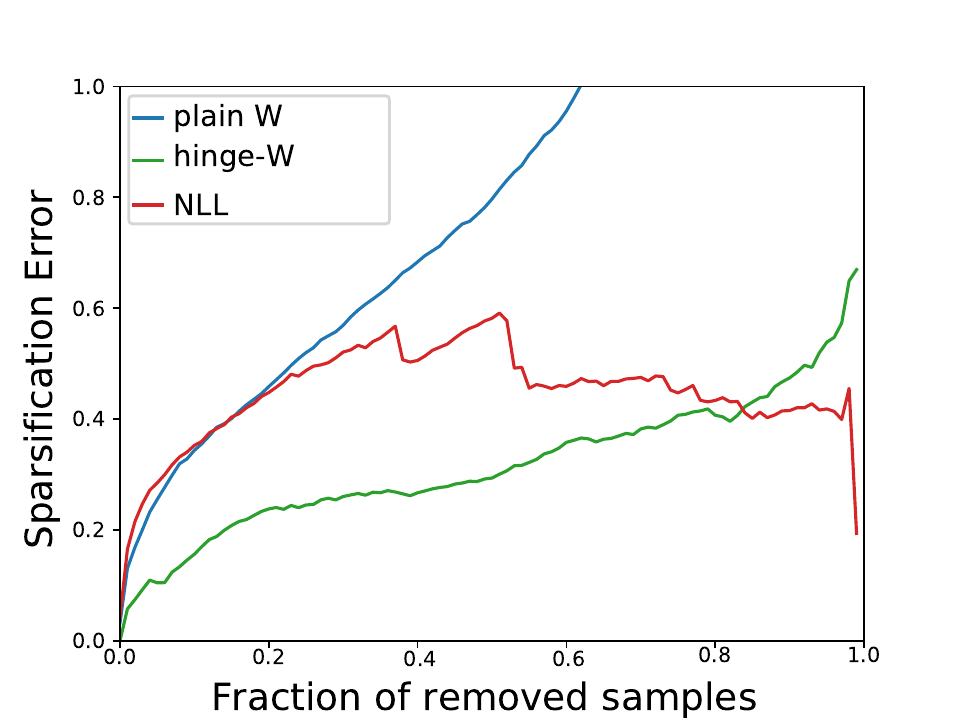} &
\includegraphics[height=4.5cm,trim = 35 0 30 35,clip]{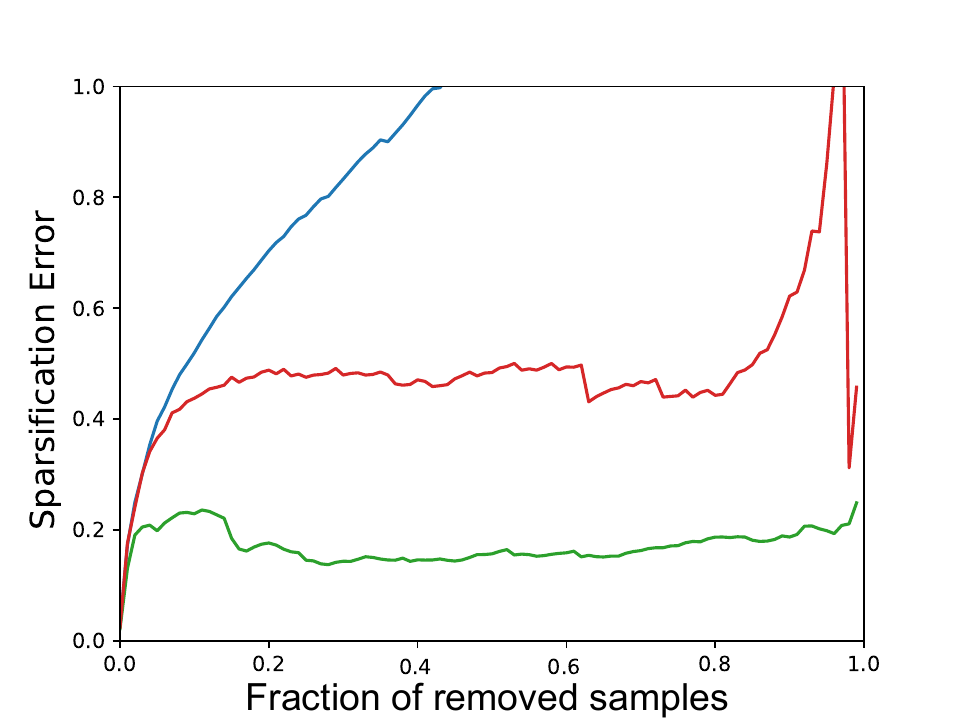} \\
(a) & (b)
\end{tabular}
}
\vspace{-3mm}
\caption{Sparsification error curves for the HLW task (lower is better). (a) $\alpha$ entropy and absolute error as the oracle. (b) Same setting for $\rho$ entropy. See Table \ref{tab:HLW} for AUSE.}
\label{fig:sparsification}
\vspace{-4mm}
\end{figure}

\begin{figure}[t]
\centering
\resizebox{0.97\linewidth}{!}{
\begin{tabular}{@{}c@{}c@{}c@{}c@{}}
\includegraphics[height=3.4cm,trim = 210 190 170 190,clip]{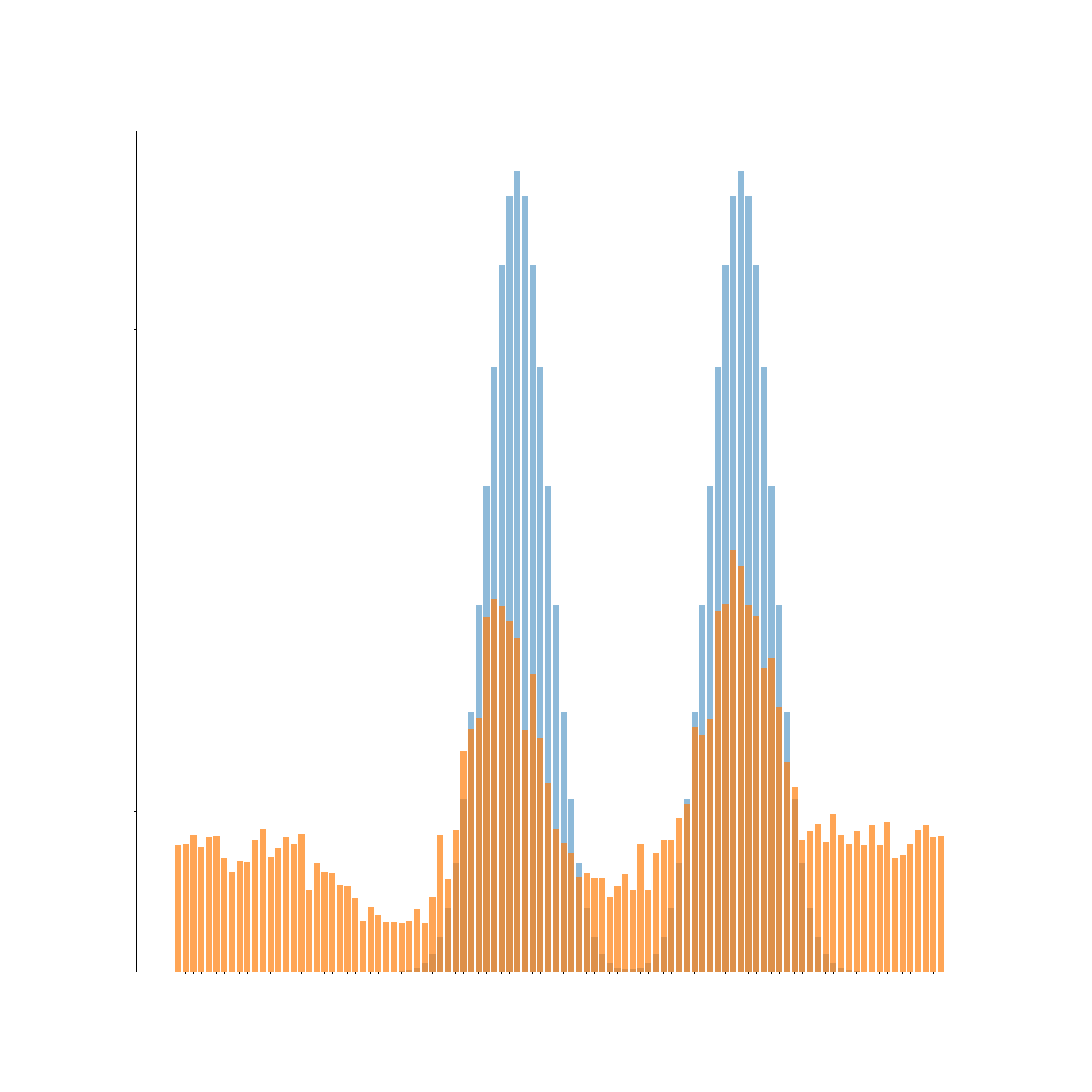} &
\includegraphics[height=3.4cm,trim = 160 190 170 190,clip]{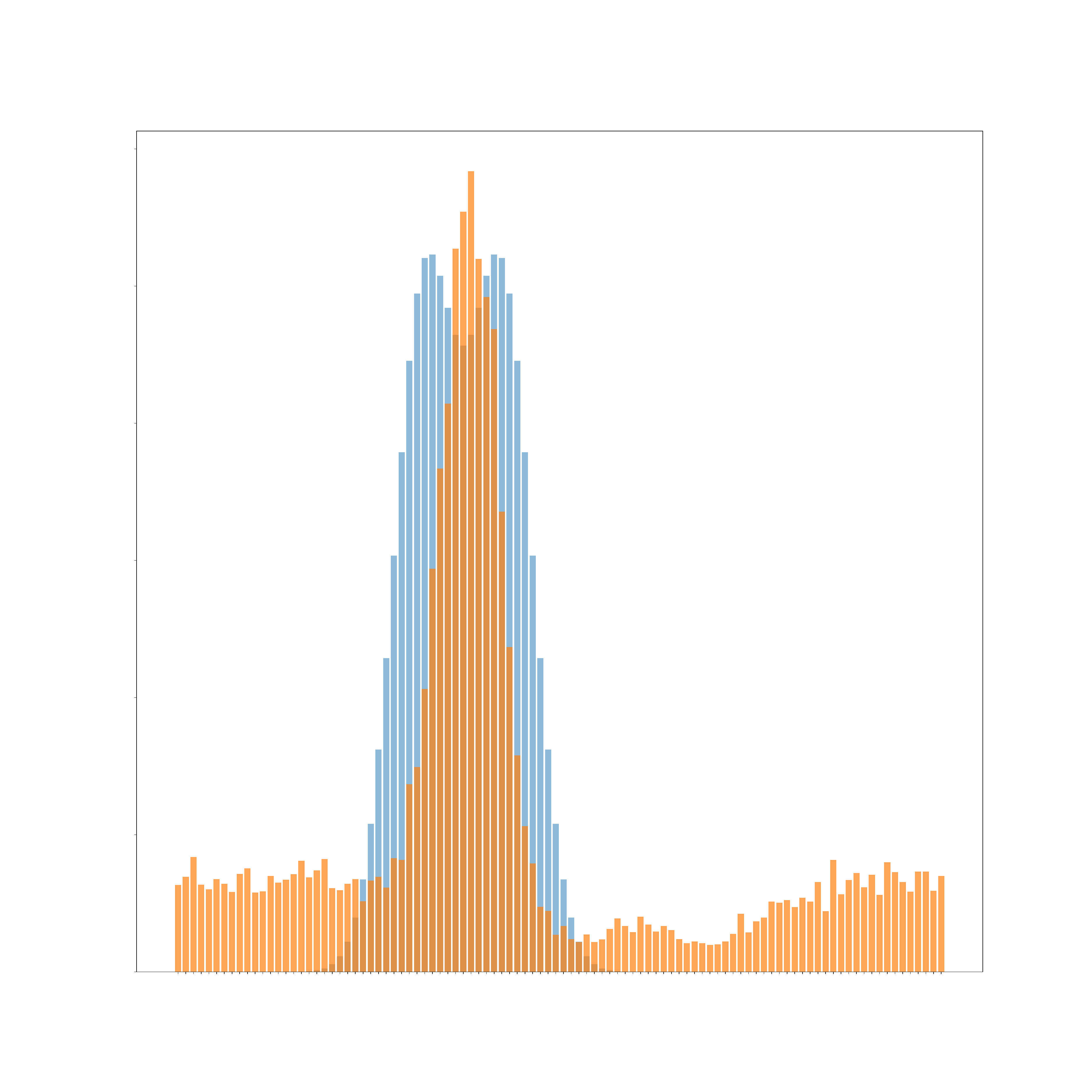} &
\includegraphics[height=3.4cm,trim = 160 190 170 190,clip]{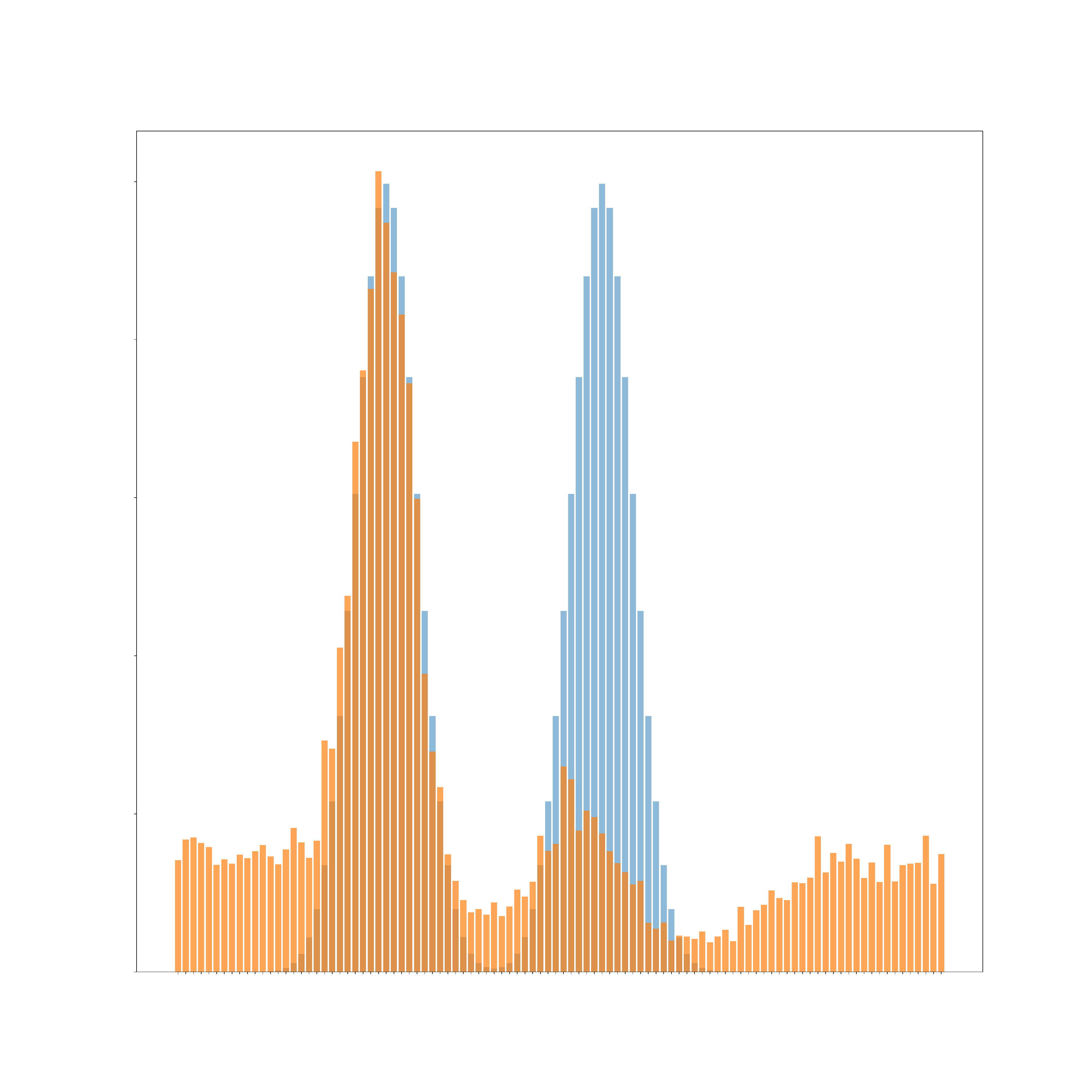} &
\includegraphics[height=3.4cm,trim = 160 190 170 190,clip]{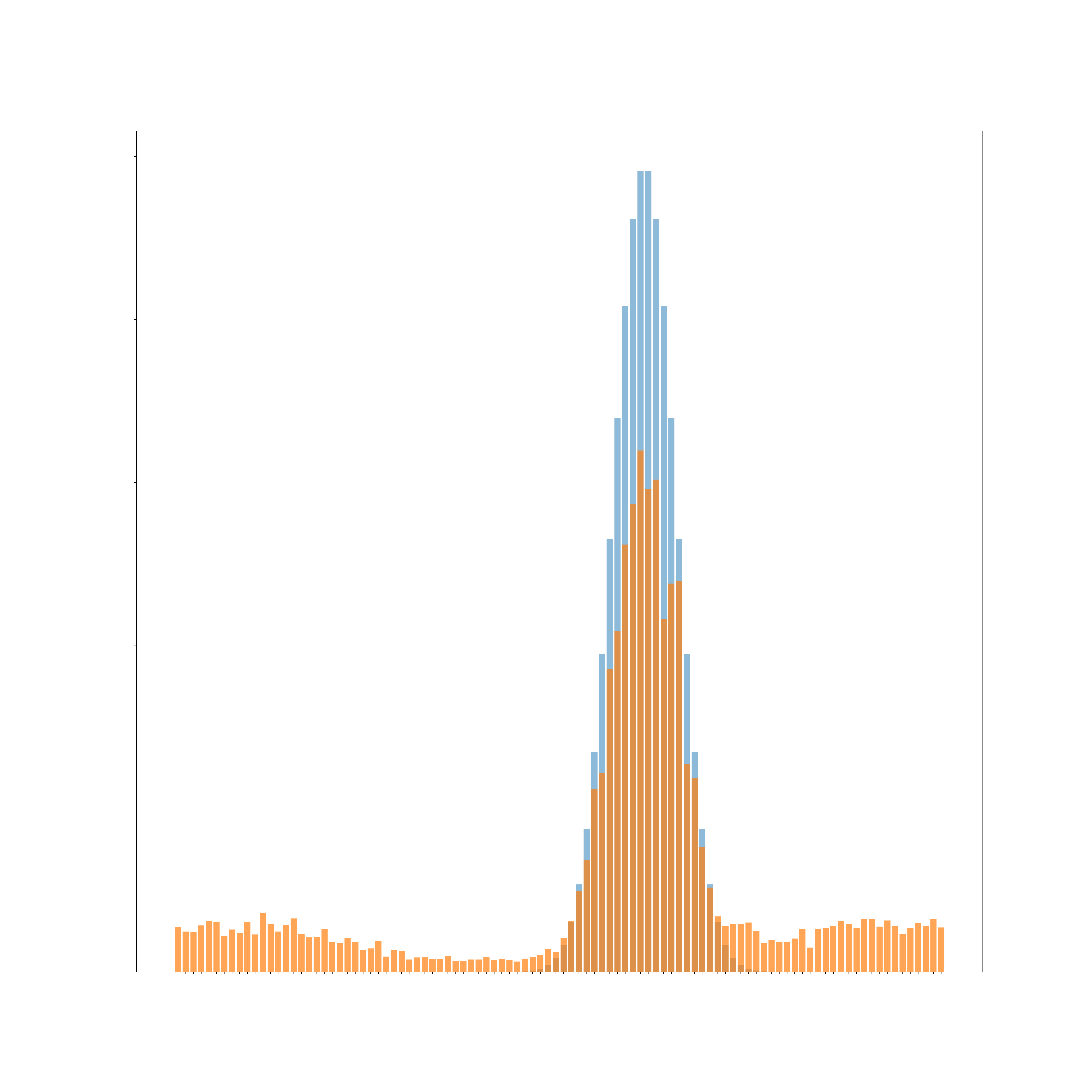}\\
(a) & (b) & (c) & (d)
\end{tabular}
}
\vspace{-3mm}
\caption{Density prediction for $\alpha$ with a model trained using hinge-$W_1$ with  $\gamma_W=0.01$, and inference on both test sets. Note: Only unimodal ground truth was used during training. Blue shows Gaussian-smoothed ground truth, and orange shows predicted densities. (a) and (c) show examples of where two output peaks overlap the ground truth; (b) shows that the model cannot distinguish two peaks if they are too close; (d) shows the model working well with unimodal ground truth.}
\label{fig:density_aleatoric}
\vspace{-5mm}
\end{figure}

\parsection{Results}:
Fig.~\ref{fig:sparsification} and Tab.~\ref{tab:HLW} show quantitative results for different configurations of our method, compared to the baseline implementation of the NLL loss \cite{workman2016hlw}. 
Using plain $W_1$ matches the baseline in terms of the AUC.
However, it causes an even more severe overconfidence problem compared to NLL, as shown by the large AUSE score. 
Hinge-$W_1$ beats the baseline by a large margin in terms of AUSE for $\alpha$ and $\rho$ yet only shows a minor decrease in AUC.  
The sparsification error plots in Fig.~\ref{fig:sparsification} indicates that NLL often leads to overconfident predictions for $\rho$ with small error as seen from the huge peak in the sparsification error curve on the right.
By contrast, hinge-$W_1$ can mitigate such overconfidence. 
In Tab.~\ref{tab:HLW} we report the results for an ensemble of $5$ NLL trained networks. As can be seen, the ensemble obtains the best AUC and slightly improves the $\alpha$ AUSE, however at the price of $5\times$ more computation.
The ensemble is actually complementary to our approach, and ensembles could potentially be combined with RbC, if a larger compute budget is available.

\begin{table*}[t]
\footnotesize
  \caption{Ablation study on the synthetic dataset. MM GND means training with multimodal ground truth (typically unavailable in real-world application); for all other results we use unimodal ground truth.
  Standard error is computed over five randomly initialized models.}
  \label{tab:synthetic}
  \vspace{-3mm}
  \centering
  \begin{tabular}{lllllllll}
  \toprule
    Loss     & AUC $\uparrow$ & $\alpha$ AUSE $\downarrow$ & $\rho$ AUSE$\downarrow$ & $\alpha$ CRPS $\downarrow$ & $\rho$ CRPS $\downarrow$\\
    \hline
    Plain \textit{$W_1$} 
    with MM GND & 46.88{\tiny $\pm$ 0.23}& 19.89{\tiny $\pm$ 1.29} & 19.49{\tiny $\pm$ 0.95} & 4.71{\tiny $\pm$0.02} & 5.20{\tiny $\pm$0.02}\\
    \hline
    Plain \textit{$W_1$} 
     & 46.36{\tiny $\pm$ 0.17} & 90.49{\tiny $\pm$3.58} & 69.67 {\tiny $\pm$ 3.83} & 8.72{\tiny $\pm$ 0.05} & 9.42{\tiny $\pm$ 0.05}\\
    \textit{hinge-$W_1$} $\gamma_W=0.005$ & 48.52{\tiny $\pm$0.12} &70.17{\tiny $\pm$2.36}&53.26{\tiny $\pm$2.62}&8.02{\tiny $\pm$0.04} & 8.42{\tiny $\pm$0.09}\\
    \textit{hinge-$W_1$} $\gamma_W=0.01$    & \textbf{49.00}{\tiny $\pm$0.09} & 63.47{\tiny $\pm$3.17} & 52.52 {\tiny $\pm$ 2.83} & \textbf{6.80}{\tiny $\pm$0.14} &\textbf{7.04}{\tiny $\pm$0.03}   \\
    \textit{hinge-$W_1$} $\gamma_W=0.015$ & 39.80{\tiny $\pm$0.14} & \textbf{21.97}{\tiny $\pm$1.54} & \textbf{28.38}{\tiny $\pm$5.15} & 9.52{\tiny $\pm$0.19} & 9.27{\tiny $\pm$0.13} \\
    \bottomrule
  \end{tabular}
 \vspace{-1mm}
\end{table*}

\begin{table*}[t]
\footnotesize
  \setlength{\tabcolsep}{5pt}
  \caption{Stereo disparity results on Scene Flow. Regression performance in terms of EPE, 1PE, and 3PE, and uncertainty evaluation in terms of entropy-based AUSE. MM denotes multimodal training with $k=5$, and standard error is reported over five runs.}
  \label{tab:disparity_all}
  \vspace{-3mm}
  \centering
  \begin{tabular}{lllllllllll}
  \toprule
  \multirow{2}{*}{Setting} & \multirow{2}{*}{Loss} & \multicolumn{4}{c}{All pixels} & \multicolumn{4}{c}{Edge pixels} \\
         ~&~& EPE $\downarrow$ & 1PE $\downarrow$ & 3PE $\downarrow$ & AUSE $\downarrow$ &EPE $\downarrow$ & 1PE $\downarrow$ & 3PE $\downarrow$ & AUSE $\downarrow$ \\ 
    \cmidrule(lr){1-6} \cmidrule(lr){7-10}
    Softmax & Plain \textit{$W_1$} \cite{div2020wstereo} & \textbf{0.98}{\tiny $\pm$0.01} & \textbf{9.44}{\tiny $\pm$0.06} & \textbf{4.04}{\tiny $\pm$0.03} & 19.4{\tiny $\pm$0.37} & \textbf{3.05}{\tiny $\pm$0.03} & \textbf{17.4}{\tiny $\pm$0.12} & \textbf{10.1}{\tiny $\pm$0.10} & 27.5{\tiny $\pm$0.70} \\ 
    Softmax & \textit{hinge-$W_1$}, $\gamma_W=0.01$ (Ours) & 0.99{\tiny $\pm$0.01} & 9.62{\tiny $\pm$0.06} & 4.08{\tiny $\pm$0.03} & \textbf{18.7}{\tiny $\pm$0.43} & \textbf{3.05}{\tiny $\pm$0.03} & 17.6{\tiny $\pm$0.12} & \textbf{10.1}{\tiny $\pm$0.06} & \textbf{26.4}{\tiny $\pm$0.31} \\ 
    \cmidrule(lr){1-6} \cmidrule(lr){7-10}
    Softplus & Plain \textit{$W_1$} \cite{div2020wstereo} & 1.00{\tiny $\pm$0.01} & 9.74{\tiny $\pm$0.07} & 4.12{\tiny $\pm$0.03} & 18.1{\tiny $\pm$0.89} & 3.05{\tiny $\pm$0.01} & 17.5{\tiny $\pm$0.09} & 10.1{\tiny $\pm$0.07} & 27.2{\tiny $\pm$1.64} \\ 
    Softplus & \textit{hinge-$W_1$}, $\gamma_W=0.0025$ (Ours) & 0.97{\tiny $\pm$0.02} & 9.35{\tiny $\pm$0.16} & 3.97{\tiny $\pm$0.06} & 16.5{\tiny $\pm$0.55} & \textbf{2.98}{\tiny $\pm$0.03} & \textbf{17.1}{\tiny $\pm$0.16} & \textbf{9.80}{\tiny $\pm$0.07} & 23.6{\tiny $\pm$0.63} \\
    Softplus & \textit{hinge-$W_1$}, $\gamma_W=0.005$ (Ours) & \textbf{0.96}{\tiny $\pm$0.01} & \textbf{9.31}{\tiny $\pm$0.05} & \textbf{3.96}{\tiny $\pm$0.03} & 16.0{\tiny $\pm$0.39} & 3.00{\tiny $\pm$0.03} & \textbf{17.1}{\tiny $\pm$0.11} & 9.84{\tiny $\pm$0.12} & 23.5{\tiny $\pm$0.58} \\
    Softplus & \textit{hinge-$W_1$}, $\gamma_W=0.01$ (Ours) & 0.98{\tiny $\pm$0.01} & 9.48{\tiny $\pm$0.06} & 4.05{\tiny $\pm$0.03} & 16.4{\tiny $\pm$0.48} & 3.04{\tiny $\pm$0.02} & 17.2{\tiny $\pm$0.16} & 9.99{\tiny $\pm$0.10} & 23.1{\tiny $\pm$0.77} \\ 
    Softplus & \textit{hinge-$W_1$}, $\gamma_W=0.015$ (Ours) & 0.97{\tiny $\pm$0.02} & 9.38{\tiny $\pm$0.13} & 3.98{\tiny $\pm$0.06} & 16.1{\tiny $\pm$0.60} & 3.01{\tiny $\pm$0.03} & 17.2{\tiny $\pm$0.21} & 9.88{\tiny $\pm$0.11} & 22.9{\tiny $\pm$0.80} \\
    Softplus & \textit{hinge-$W_1$}, $\gamma_W=0.02$ (Ours) & 0.98{\tiny $\pm$0.01} & 9.46{\tiny $\pm$0.09} & 4.04{\tiny $\pm$0.03} & \textbf{15.5}{\tiny $\pm$0.19} & 3.03{\tiny $\pm$0.03} & 17.2{\tiny $\pm$0.13} & 9.94{\tiny $\pm$0.09} & 21.8{\tiny $\pm$0.37} \\
    Softplus & \textit{hinge-$W_1$}, $\gamma_W=0.04$ (Ours) & 1.02{\tiny $\pm$0.01} & 9.75{\tiny $\pm$0.09} & 4.17{\tiny $\pm$0.05} & \textbf{15.1}{\tiny $\pm$0.26} & 3.13{\tiny $\pm$0.02} & 17.7{\tiny $\pm$0.14} & 10.3{\tiny $\pm$0.08} & \textbf{21.3}{\tiny $\pm$0.49} \\
    \cmidrule(lr){1-6} \cmidrule(lr){7-10}
    Softplus, MM & Plain \textit{$W_1$} \cite{div2020wstereo} & 1.00{\tiny $\pm$0.03} & 9.61{\tiny $\pm$0.25} & 4.15{\tiny $\pm$0.16} & 14.1{\tiny $\pm$1.46} & 3.15{\tiny $\pm$0.11} & 17.59{\tiny $\pm$0.38} & 10.3{\tiny $\pm$0.26} & 19.8{\tiny $\pm$2.08} \\ 
    Softplus, MM & \textit{hinge-$W_1$}, $\gamma_W=0.0075$ (Ours) & \textbf{0.96}{\tiny $\pm$0.01} & \textbf{9.27}{\tiny $\pm$0.11} & \textbf{3.94}{\tiny $\pm$0.03} & 13.0{\tiny $\pm$0.32}& \textbf{3.00}{\tiny $\pm$0.03} & \textbf{17.00}{\tiny $\pm$0.17} & \textbf{9.79}{\tiny $\pm$0.13} & 17.2{\tiny $\pm$0.01} \\ 
    Softplus, MM & \textit{hinge-$W_1$}, $\gamma_W=0.01$ (Ours) & 0.97{\tiny $\pm$0.03} & 9.40{\tiny $\pm$0.21} & 4.01{\tiny $\pm$0.08} & \textbf{12.6}{\tiny $\pm$0.01} & 3.04{\tiny $\pm$0.03} & 17.20{\tiny $\pm$0.08} & 9.98{\tiny $\pm$0.03} & \textbf{16.3}{\tiny $\pm$0.90} \\ 
    \bottomrule
  \end{tabular}
 \vspace{-3mm}
\end{table*}

\begin{table}[!ht]
    \footnotesize
      \caption{Test results on \textit{Horizon Lines in the Wild}. For \textit{hinge-$W_1$} and \textit{plain Wasserstein}, Gaussian smoothed ($\sigma=4$) training target apply. Both metrics are multiplied by 100.}
      \vspace{-3mm}
    \label{tab:HLW}
    \centering
    \begin{tabular}{llll}
    \toprule
        Loss & AUC $\uparrow$ & $\alpha$ AUSE $\downarrow$ & $\rho$ AUSE$\downarrow$  \\ 
    \hline
        NLL\cite{workman2016hlw}    & \textbf{64.13}{\tiny $\pm$0.04}     & 44.30{\tiny $\pm$0.86} & 51.47{\tiny $\pm$2.56} \\ 
        Ensemble NLL & \textbf{66.83}  & 39.50 & 48.00 \\
        Plain $W_1$  & 64.40{\tiny $\pm$0.22}     & 100.40{\tiny$\pm$2.76} & 153.70{\tiny $\pm$3.33} \\ 
        \textit{hinge-$W_1$},$\gamma_W$=0.0100  & 66.60{\tiny $\pm$0.11}     & 49.88{\tiny $\pm$1.73} & 78.49{\tiny $\pm$1.61} \\
        \textit{hinge-$W_1$},$\gamma_W$=0.0150  & 64.24{\tiny $\pm$0.41}     & 29.18{\tiny $\pm$0.75} & 69.41{\tiny $\pm$5.18}\\
        \textit{hinge-$W_1$},$\gamma_W$=0.0200  & 62.72{\tiny $\pm$0.14}     & \textbf{26.97}{\tiny $\pm$1.19} & \textbf{30.82}{\tiny $\pm$1.92}\\
        \textit{hinge-$W_1$},$\gamma_W$=0.0250  & 62.00{\tiny $\pm$0.09}     & 32.45{\tiny $\pm$1.36} & 31.11{\tiny $\pm$6.11}\\ 
        \bottomrule
    \end{tabular}
    \vspace{-6mm}
\end{table}

\subsection{Stereo disparity}
\label{sec:disparity}

We also study stereo disparity estimation. Given a pair of rectified stereo images, the task is to predict the disparity between their $x$-coordinates in every pixel.

\parsection{Dataset}: 
Following \cite{garg2020} we use the synthetic Scene Flow dataset \cite{MIFDB16}. 
It contains 35k training and 4k test image pairs in different domains, and forms a challenging benchmark for the disparity task. All image pairs are accompanied by unimodal ground truth disparity.

\parsection{Metrics}: To evaluate the regression performance, we measure the end-point-error (EPE), which is the absolute pixel difference between the predicted and true disparity, averaged over all pixels and images. We further measure the $k$-pixel threshold error (PE), which is the percentage of pixels where the error is more than $k$ pixels. We evaluate PE at 1-pixel and 3-pixel thresholds, denoted 1PE and 3PE. To evaluate the uncertainty estimation, we use AUSE, as described in \cref{sec:uncertainty}. 
To further invistigate, we compute all metrics on edge pixels, as these represent regions of high uncertainty, where uncertainty estimation plays a larger role.

\parsection{Implementation details}: We train a Continuous Disparity Network (CDN) \cite{div2020wstereo} on top of the PSMNet \cite{chang2018pyramid} backbone, which predicts bin center offsets in addition to the bin probabilities. This allows for continuous sub-pixel disparity predictions. We use the same settings as in \cite{div2020wstereo}, \ie we train for $10$ epochs with batch size $8$, and use a constant learning rate of $0.001$ with the Adam optimizer. The disparity is discretized into 96 bins of two pixels each, allowing for predictions in $[0, 192]$.
Furthermore, we follow \cite{div2020wstereo} in training the models in a multimodal setting, with synthetic multimodal ground truth based on neighboring pixels. In this setting, the true distribution is chosen as a Dirac mixture with disparity values based on a $k \times k$ neighborhood. The weight of the center pixel disparity is set to $\alpha=0.8$, and the rest are set with equal weight.

\parsection{Results}: \Cref{tab:disparity_all} shows the regression performance and uncertainty evaluation on the disparity task. We compare the results with softmax normalization, softplus normalization, and in multimodal training. We find that increasing the hinge $\gamma_W$ improves the uncertainty estimation, while retaining a high regression performance, in all three settings. The fact that our proposed loss yields improvements in both unimodal and multimodal settings demonstrates that it is well suited for a wide range of conditions. 
We find that both softplus normalization and synthetic multimodal disparity labels further improve the uncertainty estimation. In the results for edge pixels, the same improvements can be seen.
\label{sec:depth}

\section{Concluding remarks}
\parsection{Unavailability of multimodal groundtruth}: We generate multimodal ground truth for training on the synthetic dataset in Sec.~\ref{sec:synthetic_dataset} and on the disparity task in Sec.~\ref{sec:disparity}. 
Experiments show that MM ground truth improves uncertainty evaluation and regression for plain $W_1$. 
When we have prior knowledge about the full multimodal distribution, synthetic MM groundtruth is a way to improve uncertainty. On top of this, hinge-$W_1$ can bring further improvements.
However, in most real-world datasets, we have neither the full ground truth distribution nor the prior knowledge about it.
Our proposed hinge-$W_1$ can mitigate the absence of multimodal ground truth by reducing the penalty for secondary modes in the prediction.
A network trained with Hinge-$W_1$ and unimodal ground truth has an AUSE that is close to that from plain-$W_1$ and multimodal ground truth.

\parsection{Choice of hinge}:
All the experiments on different tasks reveal that the uncertainty in terms of AUSE keeps improving as the hinge value increases from 0 until that hinge is too large to maintain stable training.
Though not optimal, the hinge value $1/K$ ($K$ is the number of bins) already significantly improves AUSE. 
It is worth noticing that the optimal hinge is usually smaller in terms of CRPS than AUSE. This is because CRPS, as a proper scoring rule, requires sharpness of the predictive distributions. The most common use of uncertainty is to detect problematic outputs, which is more closely related to  AUSE (AUSE measures how well the uncertainty measure is able to sort the test set samples). 

We have analyzed the behaviour of the Wasserstein loss on a synthetic dataset, and shown that the absence of full ground truth distributions leads to highly overconfident unimodal predictions. We have provided a solution in the form of an added hinge, and demonstrated that this modification mitigates overconfidence, when training on datasets where full ground truth distributions are unavailable.
In the future we are interested in addressing overconfidence also in higher dimensional regression tasks, and in combining RbC with ensemble methods.



\small{\parsection{Acknowledgement}: 
This work was funded by Swedish national strategic research environment ELLIIT, grant C08, and partially by the Wallenberg Artificial Intelligence, Autonomous Systems, and Software Program (WASP) funded by the Knut and Alice Wallenberg Foundation. Computational resources were provided by the National Academic Infrastructure for Supercomputing in Sweden (NAISS).}

\clearpage
\setcounter{page}{1}
\maketitlesupplementary

\section{Introduction}
We provide in this material the contents promised in the main paper and additional results:
\cref{sec:ab_stduy} comparison between different uncertainty measures, ablation study on Gaussian smoothing and different hinge values, and additional qualitative results on HLW task (cf Sec.~3.4, Sec.~4.1, Sec.~4.2 and Sec.~4.3 in the main paper).

\section{Ablation study and additional results}
\label{sec:ab_stduy}
\subsection{Comparison between different uncertainty measures}
\label{sec:measure}
The variance, $u_\sigma$, is a common uncertainty measure for regression problems \cite{ilg2018uncertainty}:
\begin{equation}
u_\sigma(\hat{p}_y)= \sum_k k^2\hat{p}_y[k]-\left(\sum_k k\hat{p}_y[k]\right)^2
 \label{eq:std}
\end{equation}
The main disadvantage of \eqref{eq:std} is that it can easily be dominated by secondary modes that are far from the dominant mode. Another possible choice is the inverse of the maximum bin value, defined as in 
\begin{equation}
 u_M(\hat{p}_y)= \frac{1}{\hat{p}_y[k^\ast]}, k^\ast=\arg\max_k(\hat{p}_y[k])\enspace.
 \label{eq:max}
\end{equation}
 As the number of modes increases, the maximum mode will also drop, indicating larger uncertainty. In Sec.~\ref{sec:disparity} we show the effect on AUSE of these three uncertainty measures.
 
We show AUSE of varying hinge values for the stereo disparity task in \cref{tab:compasion} with three different uncertainty measures. 
All entries use the softplus activation and $L_1$ normalization as the final layer. 
Among the three measures, variance $u_\sigma$ in \cref{eq:std} achieves the smallest AUSE, which is desired for sorting predictions by uncertainty. 
We still report entropy-based AUSE in the main paper to be consistent with other tasks.
$u_M$ in \cref{eq:max} has a slightly larger AUSE but shares the same trend, that the AUSE initially optimizes towards an optimal value, and then gets worse as the hinge increases.
All three different uncertainty measures achieve the optimal AUSE at hinge, $\gamma_W=0.0075$. This shows that the improvement on AUSE from our proposed \emph{hinge-$W_1$} is robust to various uncertainty measures.
\begin{table}[t]
\footnotesize
  \caption{\textbf{Stereo disparity.} Comparison of the effect on AUSE for three different uncertainty measures, entropy, the inverse of the max bin value, and distribution variance described in  \cref{sec:measure}.}
  \label{tab:compasion}
  \centering
  \begin{tabular}{lllllllll}
  \toprule
    Settings & entropy & MAX & variance \\
    \midrule
    \textit{hinge-W}, $\gamma_W=0$ &  17.8     & 20.03 & 16.13  \\       
    \textit{hinge-W}, $\gamma_W=0.0025$ & 17.1 & 18.66 &16.00 \\
    \textit{hinge-W}, $\gamma_W=0.005$ & 15.9  & 17.40 &15.07   \\
    \textit{hinge-W}, $\gamma_W=0.0075$ & \textbf{15.9} & \textbf{17.37} & \textbf{14.87} \\
    \textit{hinge-W}, $\gamma_W=0.01$ &  17.1  & 18.80 &15.13 \\
    \textit{hinge-W}, $\gamma_W=0.0125$ & 17.5 & 19.46 &14.90 \\
    \textit{hinge-W}, $\gamma_W=0.015$ & 17.3  & 19.43 &15.33 \\
    \bottomrule
  \end{tabular}
\end{table}

Furthermore, the validity of the entropy as the scalar uncertainty measure is assessed using \textit{kernel density estimation} (KDE) plots on the two test sets. 
This is done for the entropy of the slope and offset distributions, \ie,  $u_H(\hat{p}_\alpha)$, $u_H(\hat{p}_\rho)$.
Ideally, the mode of the uncertainty measure distribution on the one-line test set should be lower and well separated from the one on the two-line test set.

\begin{figure}[t]
\centering
\begin{tabular}{@{}c@{}}
\begin{minipage}{8.5cm}
\includegraphics[width=\columnwidth,trim = 10 15 0 10,clip]{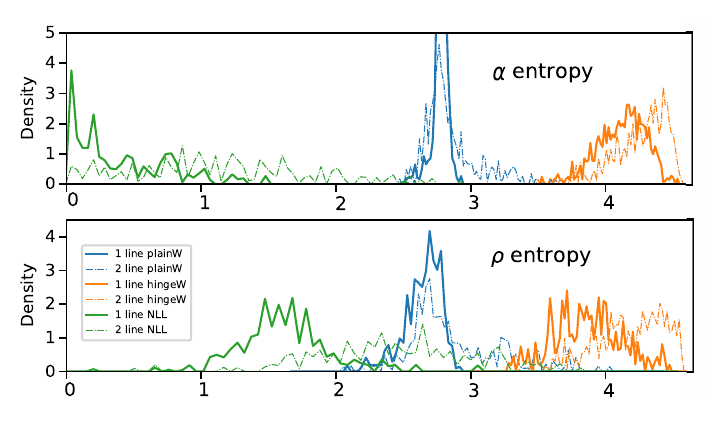}
\end{minipage}
\end{tabular}
\caption{Entropy distribution of predictions for different losses. The models are trained with unimodal annotations. Top: KDE plots for $\alpha$ entropy; Bottom: KDE plots for $\rho$ entropy; 1 line, 2 line in the legend denotes two test sets in Sec.~4.1 in the main paper. On the 2-line test set, the entropy distribution is expected to be higher than on the 1-line test set.}
\label{fig:KDE_aleatoric}
\vspace{-5mm}
\end{figure}

Fig.~\ref{fig:KDE_aleatoric} shows KDE plots for the one- and two-line test sets. 
Using the NLL loss (green) leads to a small magnitude of uncertainty for the two-line test set overlapping the peak of one-line test set.
Using the plain Wasserstein loss (blue) the network cannot distinguish ambiguous images with higher aleatoric uncertainty from others. 
The hinge-$W_1$ loss (orange) improves the separation of the modes for the two distributions.
Thus, we conclude that hinge-$W_1$ generates better aleatoric uncertainty estimates. 

\begin{table}[!ht]
    \footnotesize
      \caption{Test results on \textit{Horizon Lines in the Wild}. For \textit{hinge-$W_1$} and \textit{plain Wasserstein}, Gaussian smoothed ($\sigma=4$) training target apply. Both metrics are multiplied by 100.}
    \label{tab:HLW}
    \centering
    \begin{tabular}{llll}
    \toprule
        Loss & AUC $\uparrow$ & $\alpha$ AUSE $\downarrow$ & $\rho$ AUSE$\downarrow$  \\ 
    \hline
        NLL\cite{workman2016hlw}    & \textbf{64.13}{\tiny $\pm$0.04}     & 44.30{\tiny $\pm$0.86} & 51.47{\tiny $\pm$2.56} \\ 
        Ensemble NLL & \textbf{66.83}  & 39.50 & 48.00 \\
        Plain $W_1$  & 64.40{\tiny $\pm$0.22}     & 100.40{\tiny$\pm$2.76} & 153.70{\tiny $\pm$3.33} \\ 
        \textit{hinge-$W_1$},$\gamma_W$=0.0100  & 66.60{\tiny $\pm$0.11}     & 49.88{\tiny $\pm$1.73} & 78.49{\tiny $\pm$1.61} \\
        \textit{hinge-$W_1$},$\gamma_W$=0.0125  & 65.64{\tiny $\pm$0.14}     & 39.75{\tiny $\pm$1.40} & 64.68{\tiny $\pm$2.27}\\ 
        \textit{hinge-$W_1$},$\gamma_W$=0.0150  & 64.24{\tiny $\pm$0.41}     & 29.18{\tiny $\pm$0.75} & 69.41{\tiny $\pm$5.18}\\
        \textit{hinge-$W_1$},$\gamma_W$=0.0175  & 62.32{\tiny $\pm$0.16}     & 29.05{\tiny $\pm$1.68} & 37.34{\tiny $\pm$1.15} \\ 
        \textit{hinge-$W_1$},$\gamma_W$=0.0200  & 62.72{\tiny $\pm$0.14}     & \textbf{26.97}{\tiny $\pm$1.19} & \textbf{30.82}{\tiny $\pm$1.92}\\
        \textit{hinge-$W_1$},$\gamma_W$=0.0225  & 62.48{\tiny $\pm$0.26}     & 27.80{\tiny $\pm$0.94} & 38.83{\tiny $\pm$11.14}\\ 
        \textit{hinge-$W_1$},$\gamma_W$=0.0250  & 62.00{\tiny $\pm$0.09}     & 32.45{\tiny $\pm$1.36} & 31.11{\tiny $\pm$6.11}\\ 
        \bottomrule
    \end{tabular}
    \vspace{-6mm}
\end{table}
\subsection{Synthetic dataset: ablation on Gaussian smoothing}
In this section, we show that Gaussian smoothing is beneficial for line regression tasks. 
\cref{tab:synthetic} shows the results for training with Dirac ground truth, i.e., no Gaussian smoothing.  
As the hinge increases, both the regression performance (in terms of AUC) and uncertainty estimation (in terms of AUSE) improve until the hinge value ($\gamma_W=0.015$) is too large to maintain stable training.
Compared with Table~1 in the main paper, we notice that Dirac ground truth has much worse AUC and AUSE at hinge value ($\gamma_W=0.015$), indicating that Gaussian smoothing can help to maintain a stable training at a large hinge value.
\begin{table*}[t]
\footnotesize
  \caption{Ablation study on the synthetic dataset. For all results we use unimodal ground truth as a Dirac function.
  Standard deviation is computed over five randomly initialized models.}
  \label{tab:synthetic}
  \centering
  \begin{tabular}{lllllllll}
  \toprule
    Loss     & AUC $\uparrow$ & $\alpha$ AUSE $\downarrow$ & $\rho$ AUSE$\downarrow$ & $\alpha$ CRPS $\downarrow$ & $\rho$ CRPS $\downarrow$\\
    \hline
    Plain \textit{$W_1$} & {47.04\tiny $\pm$0.04} & {62.73\tiny$\pm$2.71} & {56.40$\pm$\tiny3.71} & {10.1\tiny $\pm$0.07} & {10.8\tiny $\pm$0.02} \\
    \textit{hinge-$W_1$} $\gamma_W=0.005$ & {47.08\tiny $\pm$1.70} & {55.51\tiny $\pm$11.35} & {40.85\tiny $\pm$14.63} & {8.55\tiny $\pm$1.79}&{9.42\tiny $\pm$1.52}\\
    \textit{hinge-$W_1$} $\gamma_W=0.01$ & {53.44\tiny $\pm$0.05} & {57.43\tiny $\pm$3.51} & {42.70\tiny $\pm$3.60} & {9.55\tiny $\pm$0.43} &{9.96\tiny $\pm$0.04}  \\
    \textit{hinge-$W_1$} $\gamma_W=0.015$ & {21.52\tiny $\pm$0.14} & {69.43\tiny $\pm$9.32} & {58.81\tiny $\pm$13.42} & {9.77\tiny $\pm$6.02} & {9.45\tiny $\pm$0.42}  \\
    \bottomrule
  \end{tabular}
\end{table*}

\subsection{HLW: ablation on different hinge}
\cref{tab:HLW} shows the ablation study on different hinge values on HLW. 
Hinge-Wasserstein with $\gamma_W=0.02$ achieves the best quality of uncertainty estimation, but it suffers a small drop in the regression performance. It is worth noting that as AUC kepdf dropping as $\gamma_W$ increases, whereas AUSE first drops and then increases. This indicates there exists an optimal value $\gamma_W^*$ on the HLW dataset. When $\gamma_W > \gamma_W^*$, it will be rather hard to train the neural network, as there will rarely be any gradients from the loss. We also argue that $\gamma_W^*$ depends on the number of bins in the regression by classification framework. E.g., there are 100 bins for the horizon line detection task, and thus, $\gamma_W=1/100$ means that hinge-Wasserstein allows a random guess.

\subsection{Stereo disparity: ablation on different hinge}
\label{sec:disparity}
We report the results of different hinge values for the stereo disparity task in \cref{tab:disparity_all}. As hinge increases, both regression performance (in terms of EPE) and uncertainty estimation (in terms of AUSE) improved for both boundary pixels and all the pixels. This shows our proposed hinge-$W_1$ improves the challenging multimodal regression.
\begin{table*}[t]
\footnotesize
  \setlength{\tabcolsep}{5pt}
  \caption{Stereo disparity results on Scene Flow. Regression performance in terms of EPE, 1PE, and 3PE, and uncertainty evaluation in terms of entropy-based AUSE. MM denotes multimodal training with $k=5$, and standard error is reported over five runs.}
  \label{tab:disparity_all}
  \centering
  \begin{tabular}{lllllllllll}
  \toprule
  \multirow{2}{*}{Setting} & \multirow{2}{*}{Loss} & \multicolumn{4}{c}{All pixels} & \multicolumn{4}{c}{Edge pixels} \\
         ~&~& EPE $\downarrow$ & 1PE $\downarrow$ & 3PE $\downarrow$ & AUSE $\downarrow$ &EPE $\downarrow$ & 1PE $\downarrow$ & 3PE $\downarrow$ & AUSE $\downarrow$ \\ 
    \cmidrule(lr){1-6} \cmidrule(lr){7-10}
    Softmax & Plain \textit{$W_1$} \cite{div2020wstereo} & \textbf{0.98}{\tiny $\pm$0.01} & \textbf{9.44}{\tiny $\pm$0.06} & \textbf{4.04}{\tiny $\pm$0.03} & 19.4{\tiny $\pm$0.37} & \textbf{3.05}{\tiny $\pm$0.03} & \textbf{17.4}{\tiny $\pm$0.12} & \textbf{10.1}{\tiny $\pm$0.10} & 27.5{\tiny $\pm$0.70} \\ 
    Softmax & \textit{hinge-$W_1$}, $\gamma_W=0.0075$ (Ours) & 1.03{\tiny $\pm$0.02} & 9.80{\tiny $\pm$0.11} & 4.19{\tiny $\pm$0.05} & \textbf{18.7}{\tiny $\pm$0.29} & 3.11{\tiny $\pm$0.01} & 17.8{\tiny $\pm$0.09} & 10.3{\tiny $\pm$0.05} & 26.7{\tiny $\pm$0.41} \\ 
    Softmax & \textit{hinge-$W_1$}, $\gamma_W=0.01$ (Ours) & 0.99{\tiny $\pm$0.01} & 9.62{\tiny $\pm$0.06} & 4.08{\tiny $\pm$0.03} & \textbf{18.7}{\tiny $\pm$0.43} & \textbf{3.05}{\tiny $\pm$0.03} & 17.6{\tiny $\pm$0.12} & \textbf{10.1}{\tiny $\pm$0.06} & \textbf{26.4}{\tiny $\pm$0.31} \\ 
    \cmidrule(lr){1-6} \cmidrule(lr){7-10}
    Softplus & Plain \textit{$W_1$} \cite{div2020wstereo} & 1.00{\tiny $\pm$0.01} & 9.74{\tiny $\pm$0.07} & 4.12{\tiny $\pm$0.03} & 18.1{\tiny $\pm$0.89} & 3.05{\tiny $\pm$0.01} & 17.5{\tiny $\pm$0.09} & 10.1{\tiny $\pm$0.07} & 27.2{\tiny $\pm$1.64} \\ 
    Softplus & \textit{hinge-$W_1$}, $\gamma_W=0.001$ (Ours) & 0.97{\tiny $\pm$0.01} & 9.48{\tiny $\pm$0.05} & 4.05{\tiny $\pm$0.03} & 17.4{\tiny $\pm$0.44} & 3.00{\tiny $\pm$0.02} & 17.2{\tiny $\pm$0.12} & 9.91{\tiny $\pm$0.05} & 26.1{\tiny $\pm$0.59} \\
    Softplus & \textit{hinge-$W_1$}, $\gamma_W=0.0025$ (Ours) & 0.97{\tiny $\pm$0.02} & 9.35{\tiny $\pm$0.16} & 3.97{\tiny $\pm$0.06} & 16.5{\tiny $\pm$0.55} & \textbf{2.98}{\tiny $\pm$0.03} & \textbf{17.1}{\tiny $\pm$0.16} & \textbf{9.80}{\tiny $\pm$0.07} & 23.6{\tiny $\pm$0.63} \\
    Softplus & \textit{hinge-$W_1$}, $\gamma_W=0.005$ (Ours) & \textbf{0.96}{\tiny $\pm$0.01} & \textbf{9.31}{\tiny $\pm$0.05} & \textbf{3.96}{\tiny $\pm$0.03} & 16.0{\tiny $\pm$0.39} & 3.00{\tiny $\pm$0.03} & \textbf{17.1}{\tiny $\pm$0.11} & 9.84{\tiny $\pm$0.12} & 23.5{\tiny $\pm$0.58} \\
    Softplus & \textit{hinge-$W_1$}, $\gamma_W=0.0075$ (Ours) & 1.00{\tiny $\pm$0.01} & 9.52{\tiny $\pm$0.06} & 4.06{\tiny $\pm$0.04} & 15.6{\tiny $\pm$0.33} & 3.07{\tiny $\pm$0.01} & 17.4{\tiny $\pm$0.09} & 10.0{\tiny $\pm$0.07} & 23.2{\tiny $\pm$0.71} \\ 
    Softplus & \textit{hinge-$W_1$}, $\gamma_W=0.01$ (Ours) & 0.98{\tiny $\pm$0.01} & 9.48{\tiny $\pm$0.06} & 4.05{\tiny $\pm$0.03} & 16.4{\tiny $\pm$0.48} & 3.04{\tiny $\pm$0.02} & 17.2{\tiny $\pm$0.16} & 9.99{\tiny $\pm$0.10} & 23.1{\tiny $\pm$0.77} \\ 
    Softplus & \textit{hinge-$W_1$}, $\gamma_W=0.0125$ (Ours) & 1.00{\tiny $\pm$0.01} & 9.60{\tiny $\pm$0.08} & 4.12{\tiny $\pm$0.05} & 15.8{\tiny $\pm$0.27} & 3.06{\tiny $\pm$0.02} & 17.4{\tiny $\pm$0.08} & 10.0{\tiny $\pm$0.06} & \textbf{21.7}{\tiny $\pm$0.50} \\
    Softplus & \textit{hinge-$W_1$}, $\gamma_W=0.015$ (Ours) & 0.97{\tiny $\pm$0.02} & 9.38{\tiny $\pm$0.13} & 3.98{\tiny $\pm$0.06} & 16.1{\tiny $\pm$0.60} & 3.01{\tiny $\pm$0.03} & 17.2{\tiny $\pm$0.21} & 9.88{\tiny $\pm$0.11} & 22.9{\tiny $\pm$0.80} \\
    Softplus & \textit{hinge-$W_1$}, $\gamma_W=0.02$ (Ours) & 0.98{\tiny $\pm$0.01} & 9.46{\tiny $\pm$0.09} & 4.04{\tiny $\pm$0.03} & \textbf{15.5}{\tiny $\pm$0.19} & 3.03{\tiny $\pm$0.03} & 17.2{\tiny $\pm$0.13} & 9.94{\tiny $\pm$0.09} & 21.8{\tiny $\pm$0.37} \\
    Softplus & \textit{hinge-$W_1$}, $\gamma_W=0.04$ (Ours) & 1.02{\tiny $\pm$0.01} & 9.75{\tiny $\pm$0.09} & 4.17{\tiny $\pm$0.05} & \textbf{15.1}{\tiny $\pm$0.26} & 3.13{\tiny $\pm$0.02} & 17.7{\tiny $\pm$0.14} & 10.3{\tiny $\pm$0.08} & \textbf{21.3}{\tiny $\pm$0.49} \\
    \cmidrule(lr){1-6} \cmidrule(lr){7-10}
    Softplus, MM & Plain \textit{$W_1$} \cite{div2020wstereo} & 1.00{\tiny $\pm$0.03} & 9.61{\tiny $\pm$0.25} & 4.15{\tiny $\pm$0.16} & 14.1{\tiny $\pm$1.46} & 3.15{\tiny $\pm$0.11} & 17.59{\tiny $\pm$0.38} & 10.3{\tiny $\pm$0.26} & 19.8{\tiny $\pm$2.08} \\ 
    Softplus, MM & \textit{hinge-$W_1$}, $\gamma_W=0.0075$ (Ours) & \textbf{0.96}{\tiny $\pm$0.01} & \textbf{9.27}{\tiny $\pm$0.11} & \textbf{3.94}{\tiny $\pm$0.03} & 13.0{\tiny $\pm$0.32}& \textbf{3.00}{\tiny $\pm$0.03} & \textbf{17.00}{\tiny $\pm$0.17} & \textbf{9.79}{\tiny $\pm$0.13} & 17.2{\tiny $\pm$0.01} \\ 
    Softplus, MM & \textit{hinge-$W_1$}, $\gamma_W=0.01$ (Ours) & 0.97{\tiny $\pm$0.03} & 9.40{\tiny $\pm$0.21} & 4.01{\tiny $\pm$0.08} & \textbf{12.6}{\tiny $\pm$0.01} & 3.04{\tiny $\pm$0.03} & 17.20{\tiny $\pm$0.08} & 9.98{\tiny $\pm$0.03} & \textbf{16.3}{\tiny $\pm$0.90} \\ 
    \bottomrule
  \end{tabular}
 \vspace{-3mm}
\end{table*}

\section{Horizon in the wild: additional qualitative results}
Fig.~\ref{fig:ambiguity_illustration} shows more examples of images and the corresponding predicted densities for $\alpha$ and $\rho$. 
The peak shapes are more clearly defined for $\alpha$ than for $\rho$. This is a general trend that we have observed, and it is also consistent with the more focused curves for alpha at the bottom of Fig. \ref{fig:ambiguity_illustration}.
\begin{figure*}[t]
\centering
\begin{tabular}{@{}ccc@{}}
\includegraphics[height=3cm,trim = 0 0 20 0,clip]{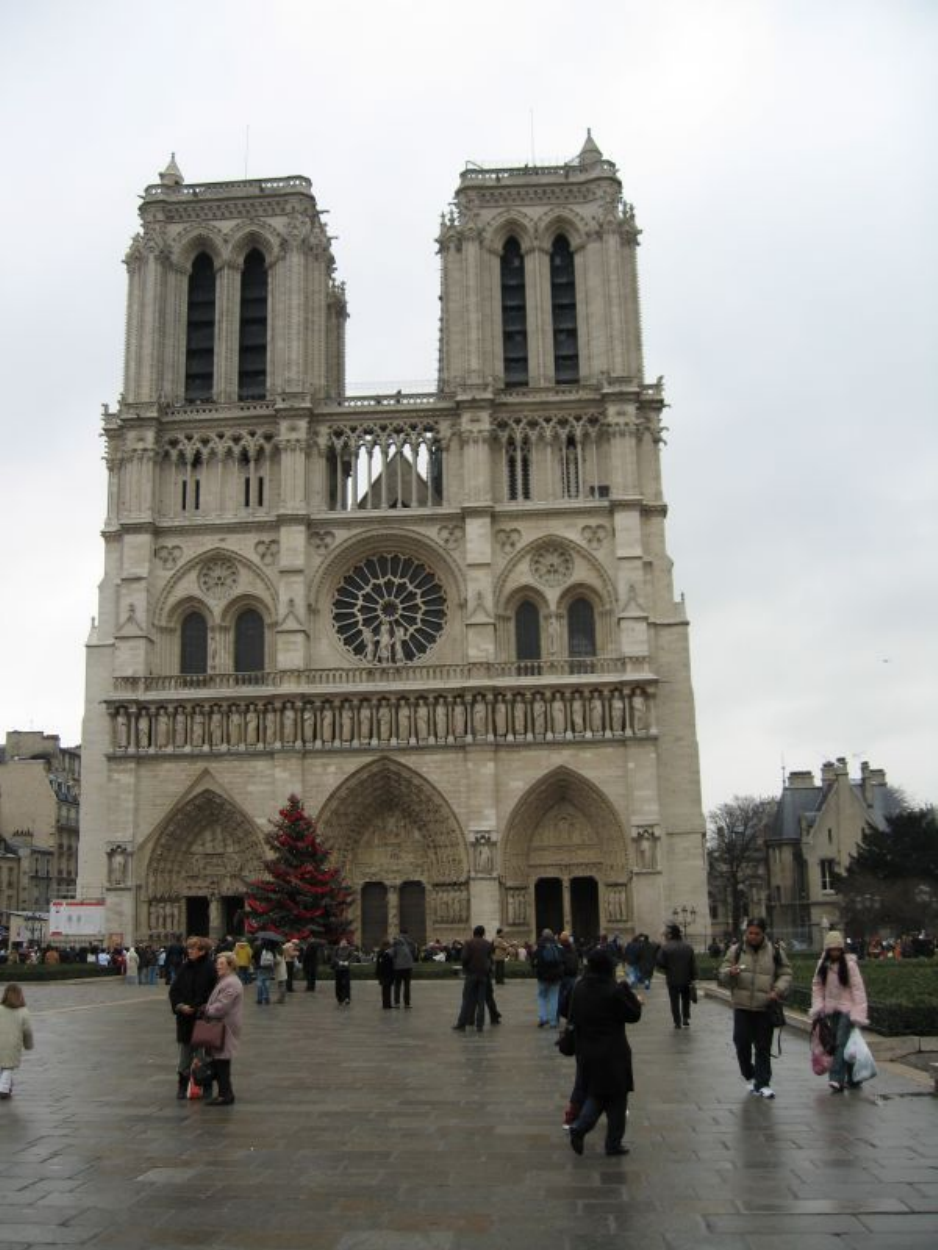} &
\includegraphics[height=3cm]{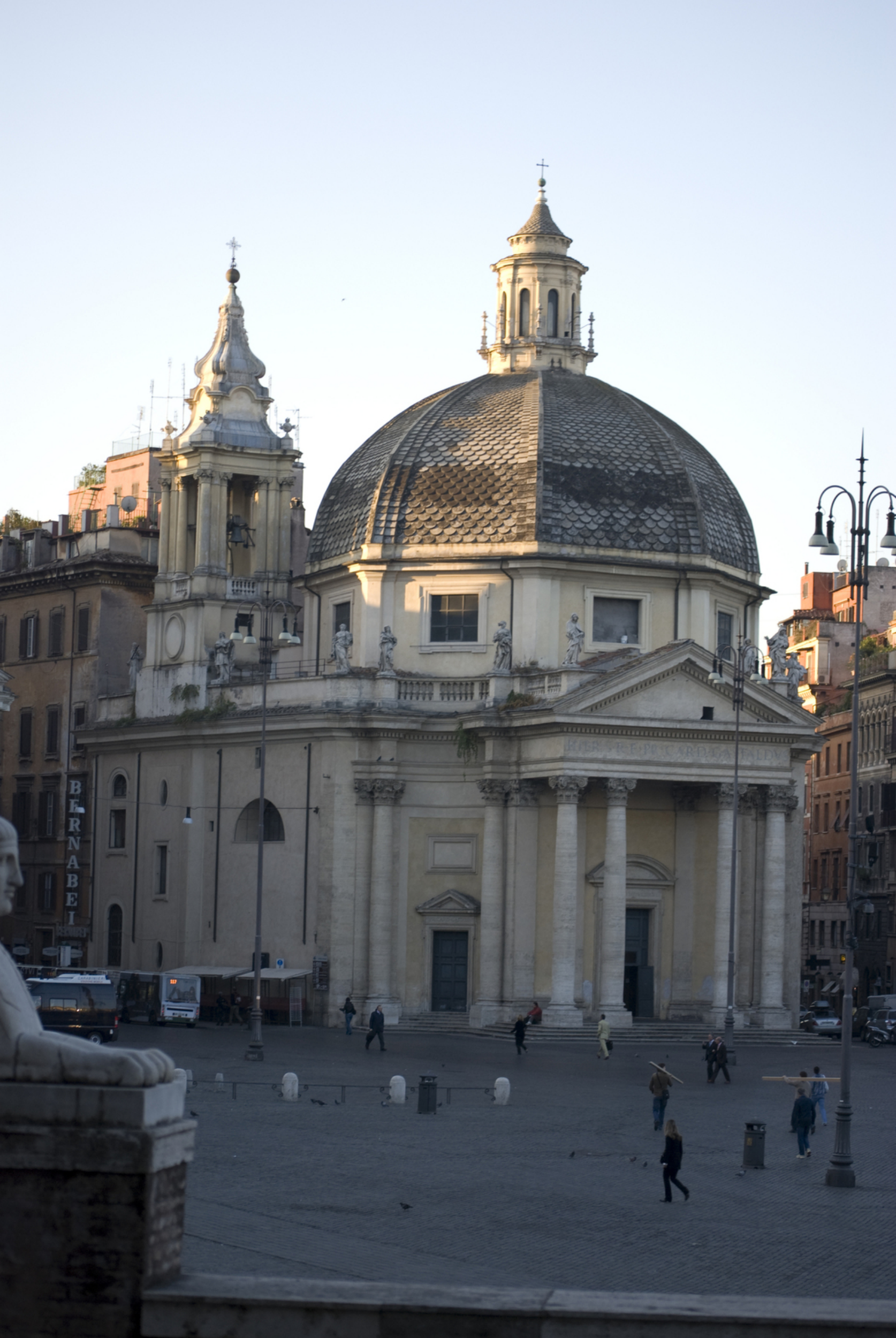} &
\includegraphics[height=3cm,trim = 0 20 20 0,clip]{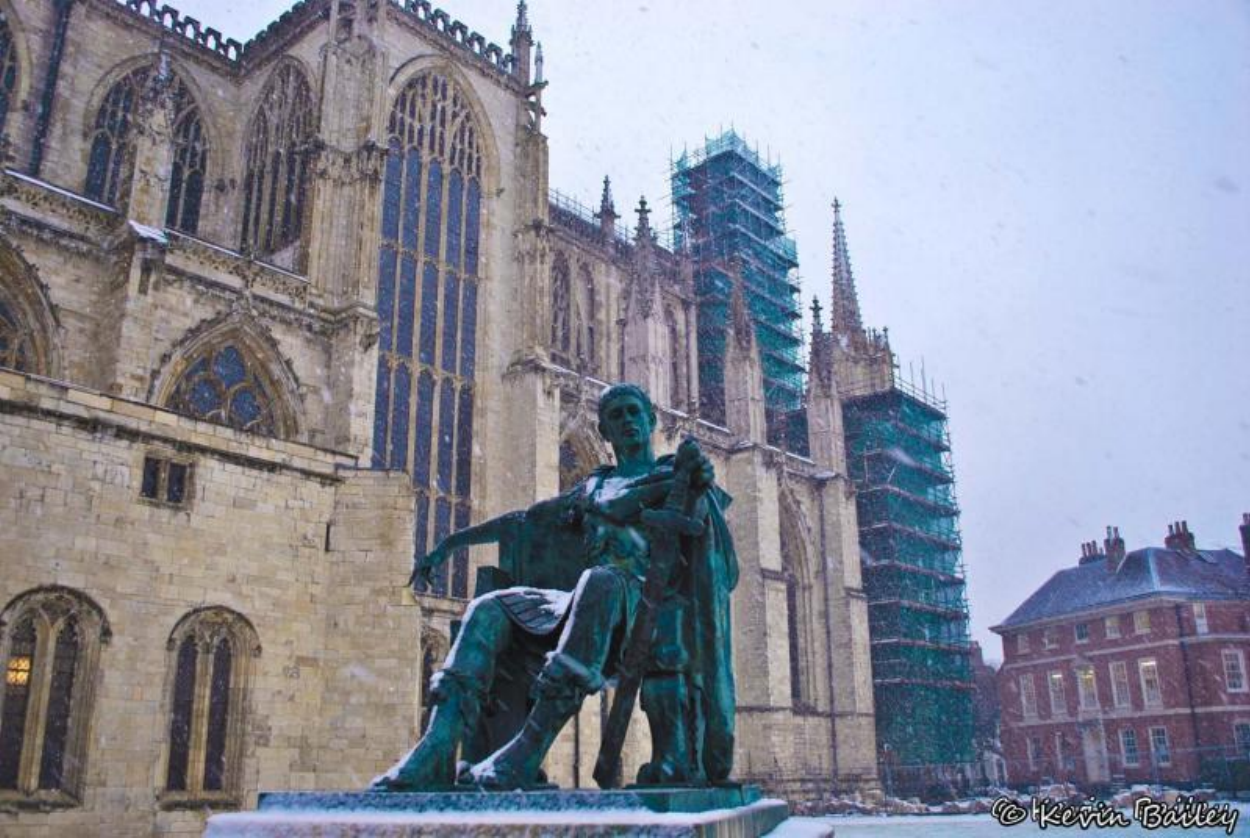} \\[-0.5ex]
\begin{tabular}{@{}lr@{}}
\includegraphics[width=1.8cm,height=1.2cm,trim = 200 190 180 210,clip]{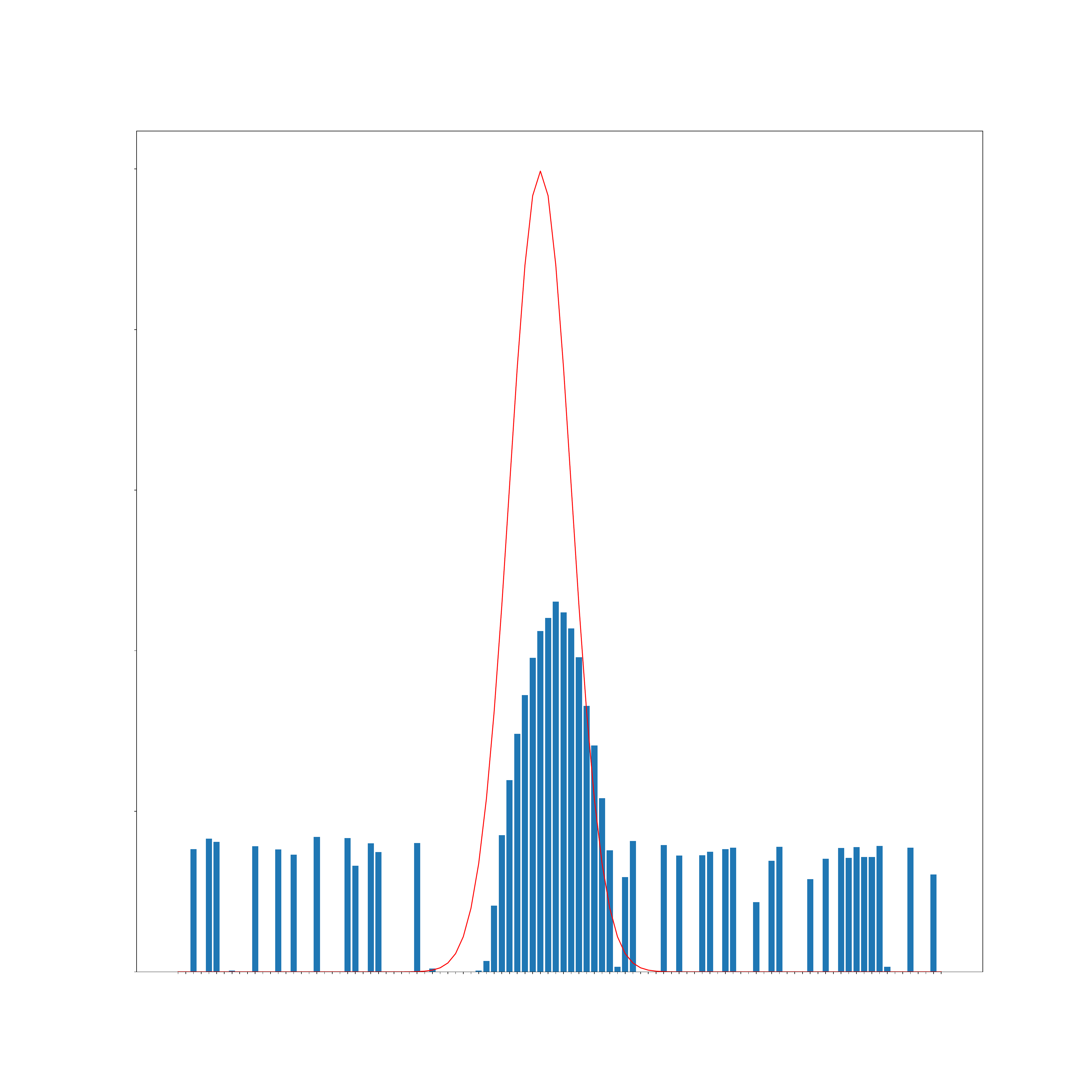} &
\includegraphics[width=1.8cm,height=1.2cm,trim = 200 190 180 210,clip]{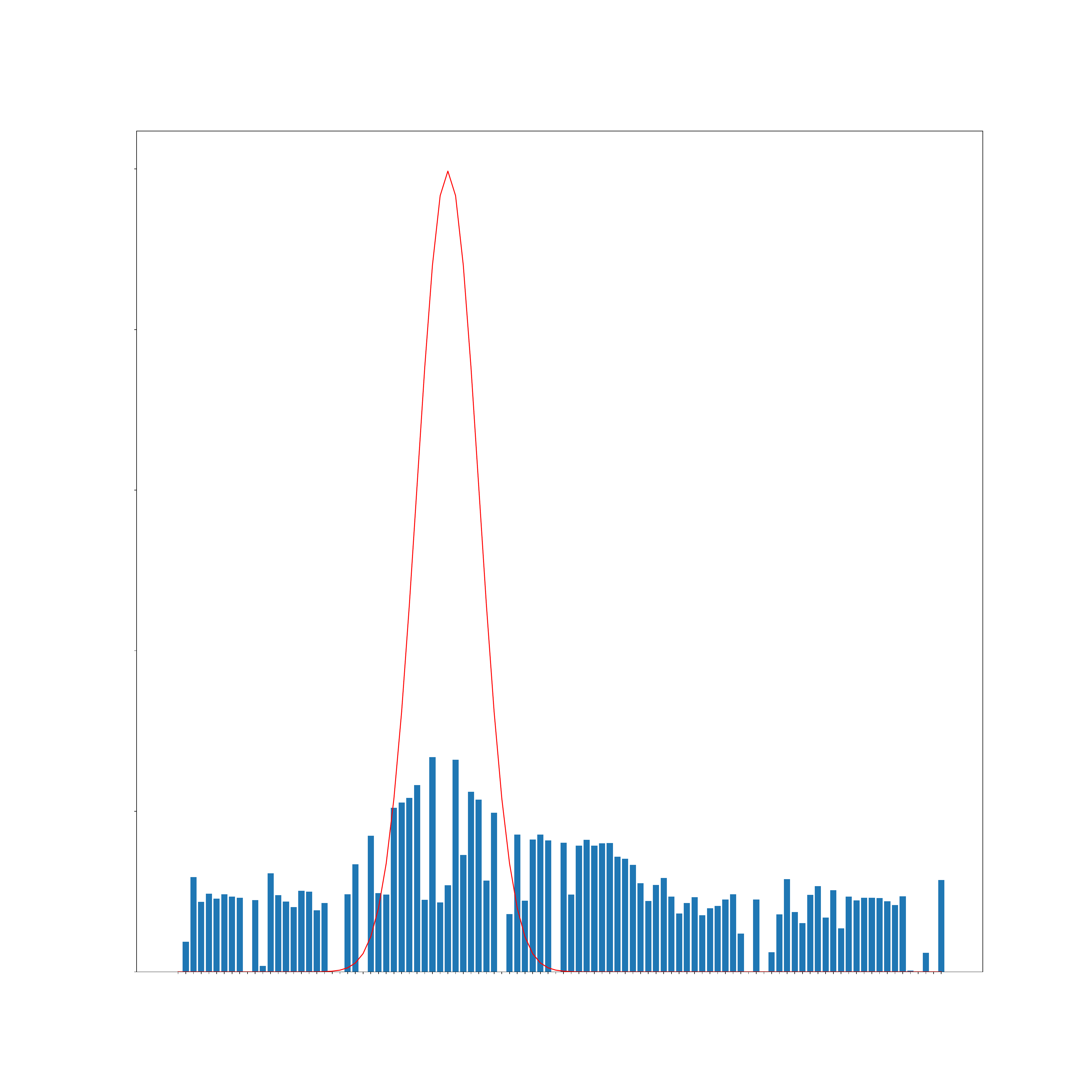}
\end{tabular} &
\begin{tabular}{@{}lr@{}}
\includegraphics[width=1.8cm,height=1.2cm,trim = 200 190 180 210,clip]{img/9_theta_img737.pdf} &
\includegraphics[width=1.8cm,height=1.2cm,trim = 200 190 180 210,clip]{img/9_rho_img737.pdf} 
\end{tabular} &

\begin{tabular}{@{}lr@{}}
\includegraphics[width=1.8cm,height=1.2cm,trim = 200 190 180 210,clip]{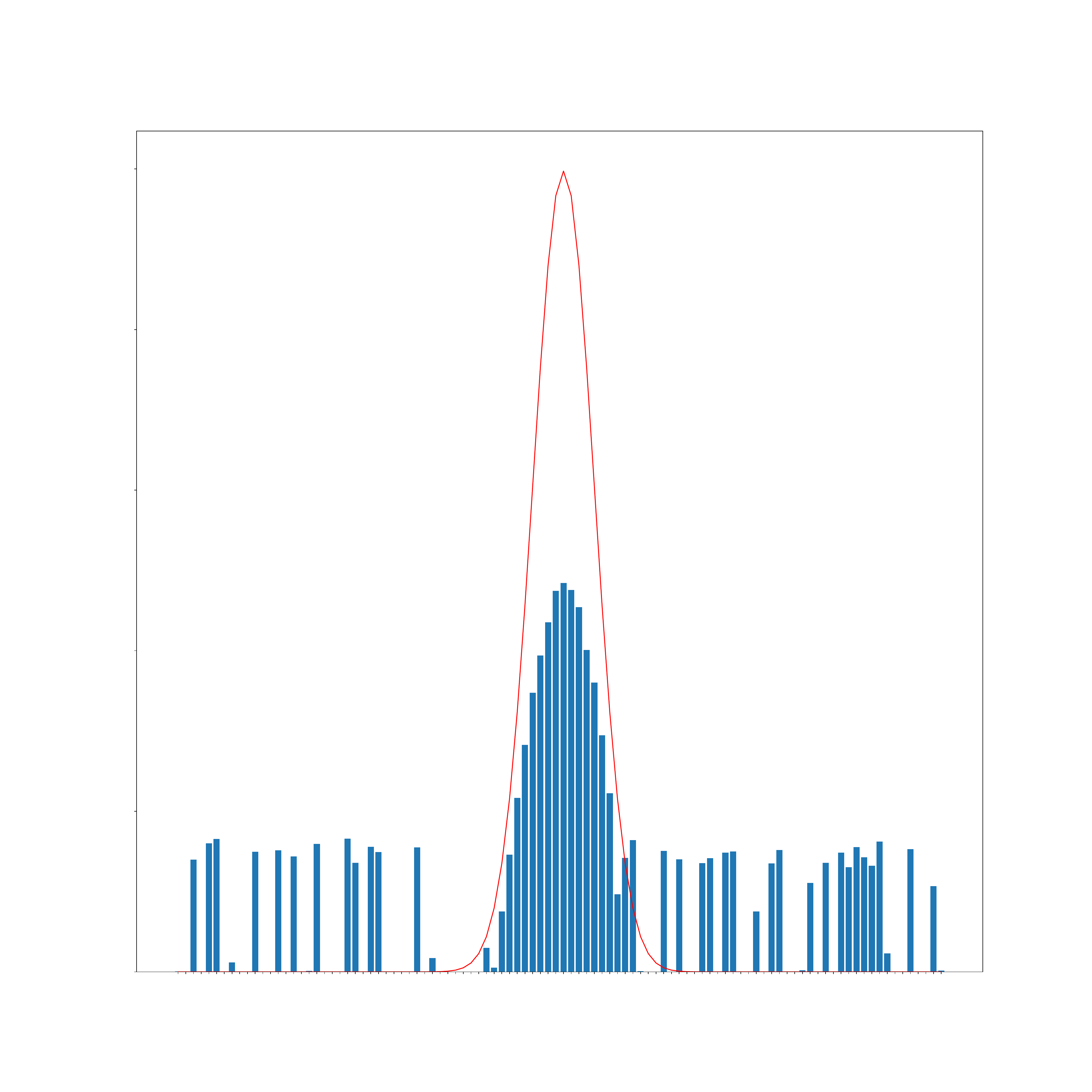} &
\includegraphics[width=1.8cm,height=1.2cm,trim = 200 190 180 210,clip]{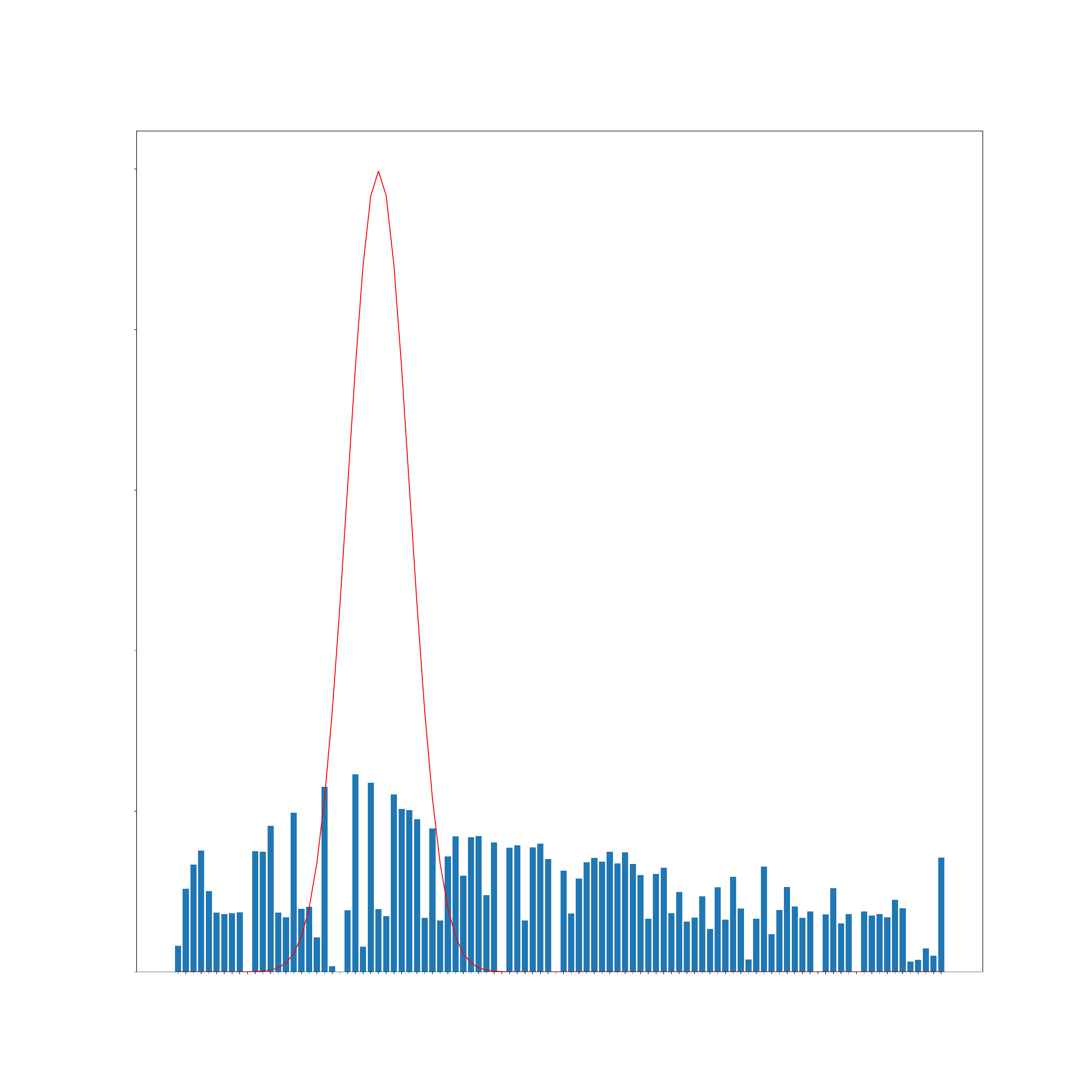}
\end{tabular}\\

\includegraphics[height=3cm,trim = 0 0 20 0,clip]{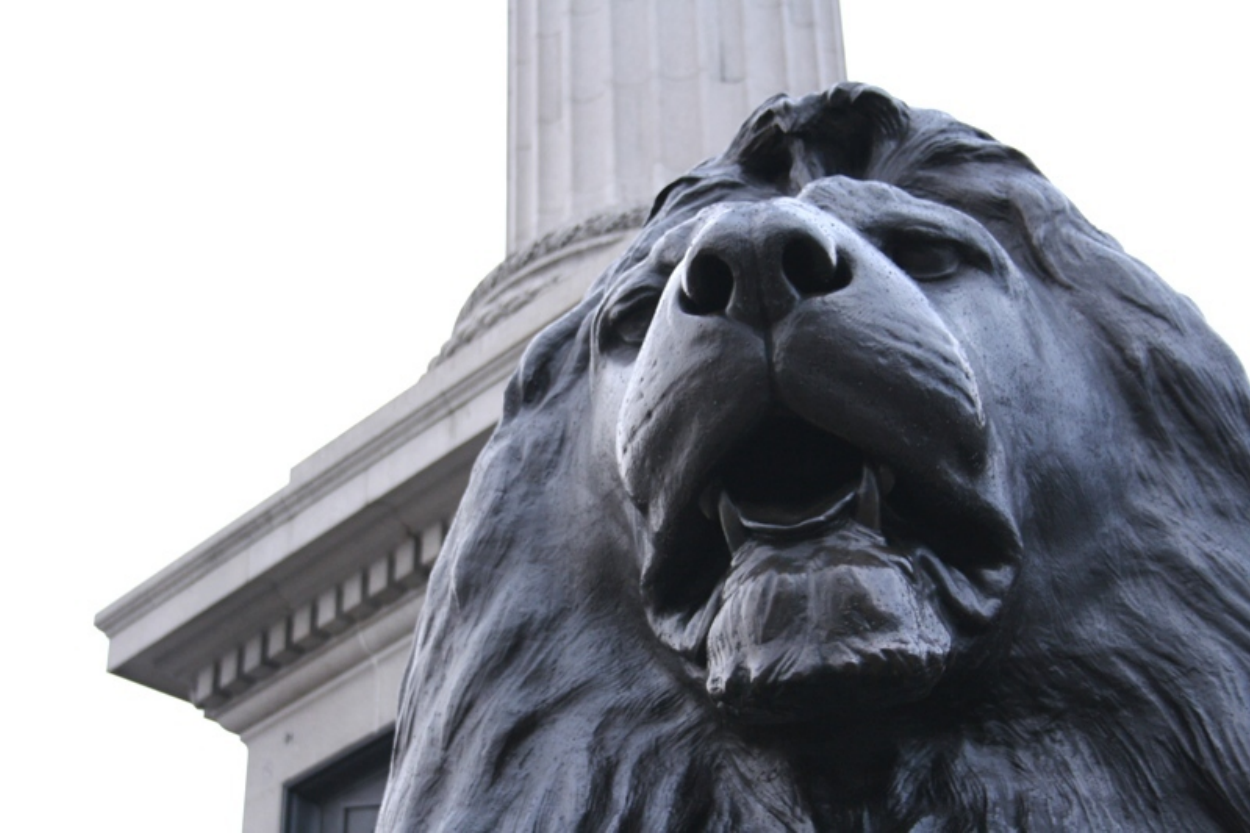} &
\includegraphics[height=3cm]{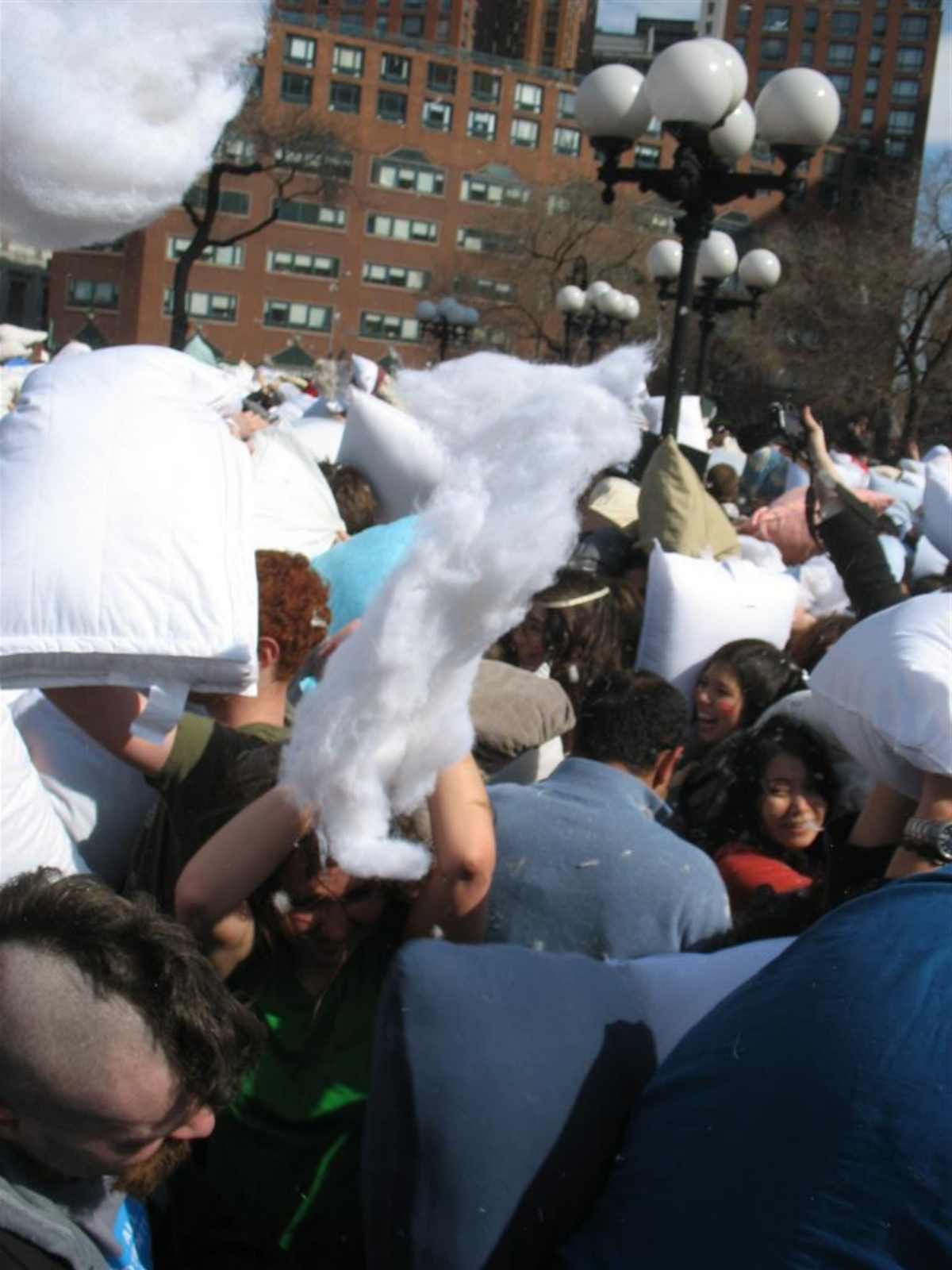} &
\includegraphics[height=3cm,trim = 0 20 20 0,clip]{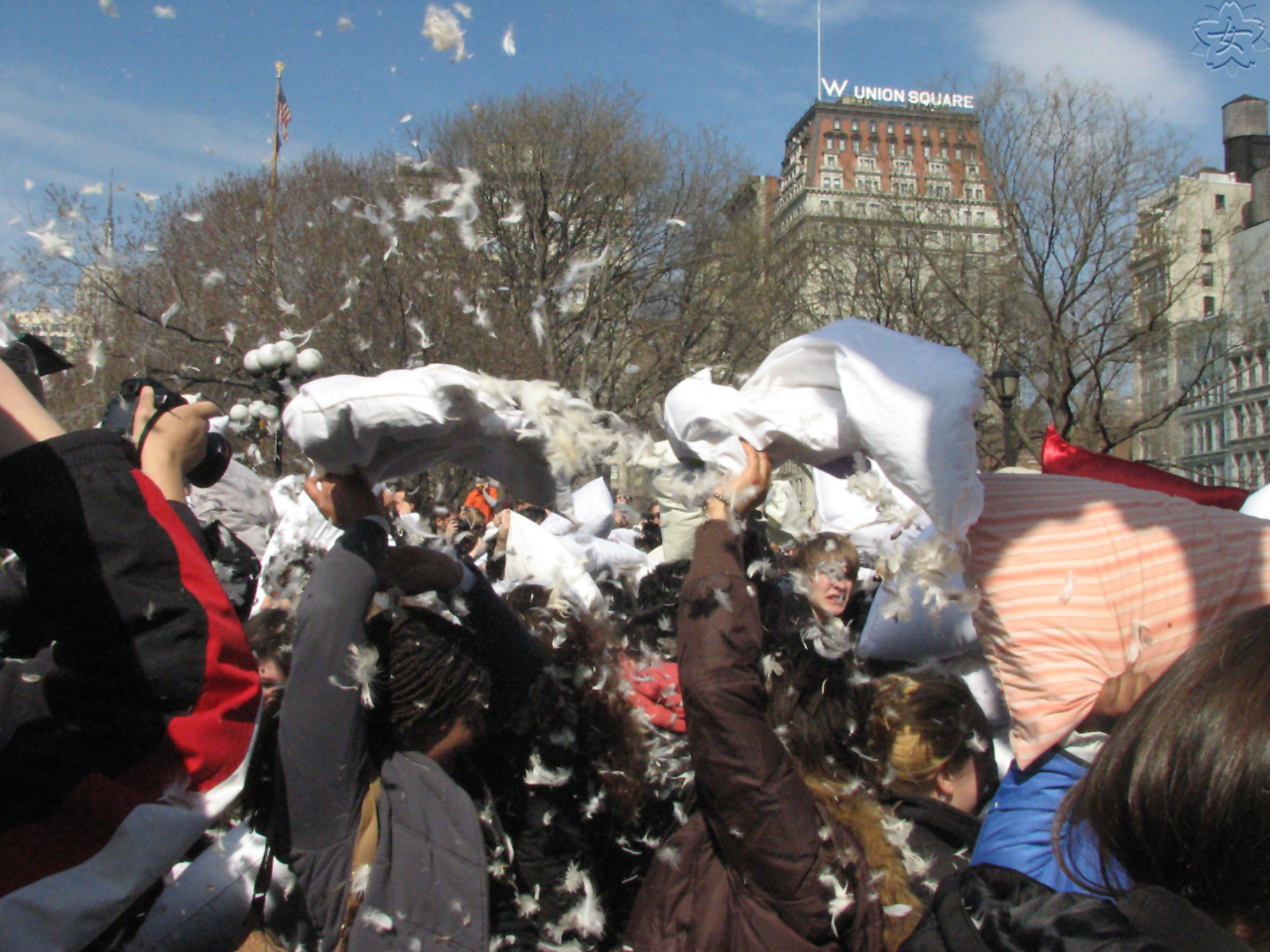} \\[-0.5ex]

\begin{tabular}{@{}lr@{}}
\includegraphics[width=1.8cm,height=1.2cm,trim = 200 190 180 210,clip]{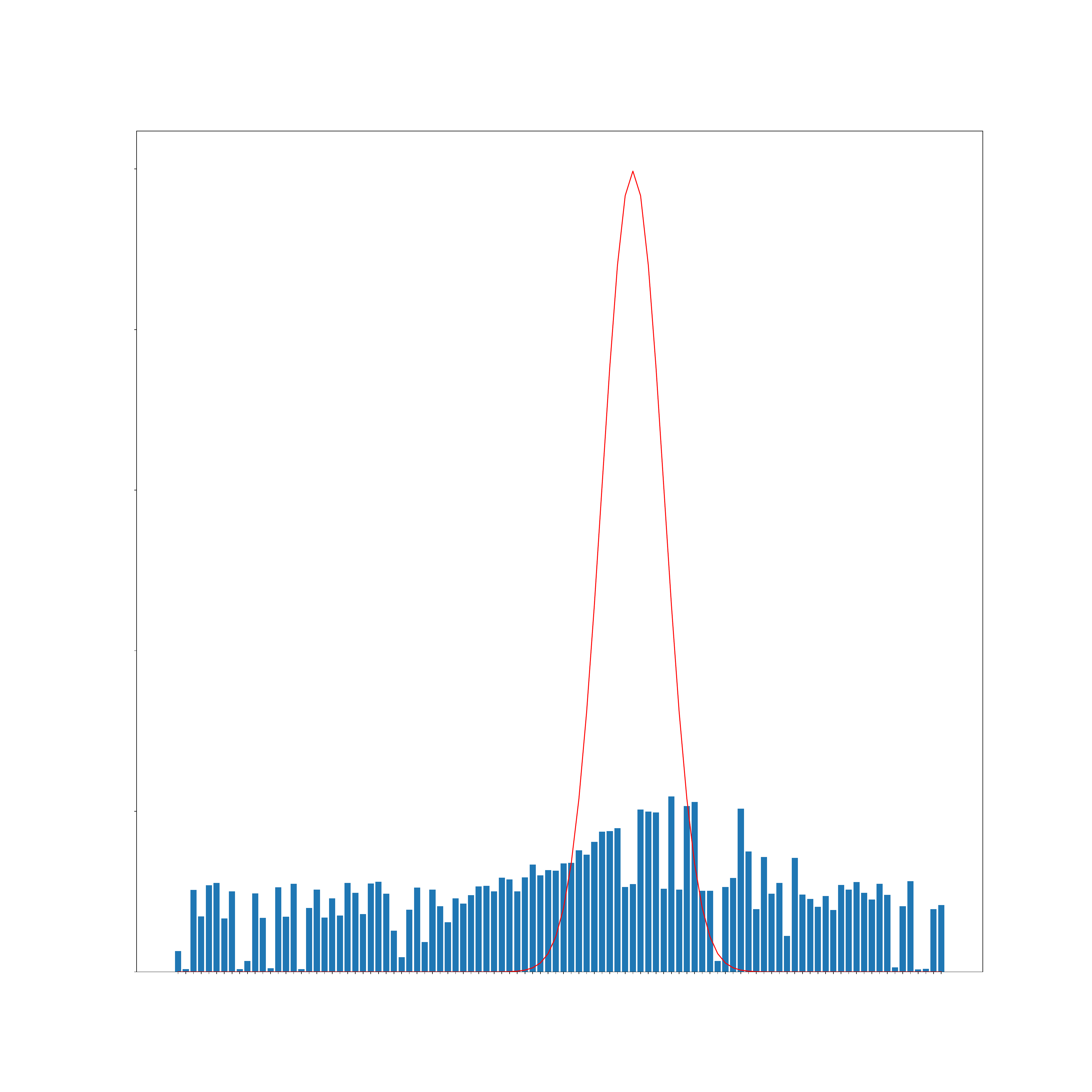} &
\includegraphics[width=1.8cm,height=1.2cm,trim = 200 190 180 210,clip]{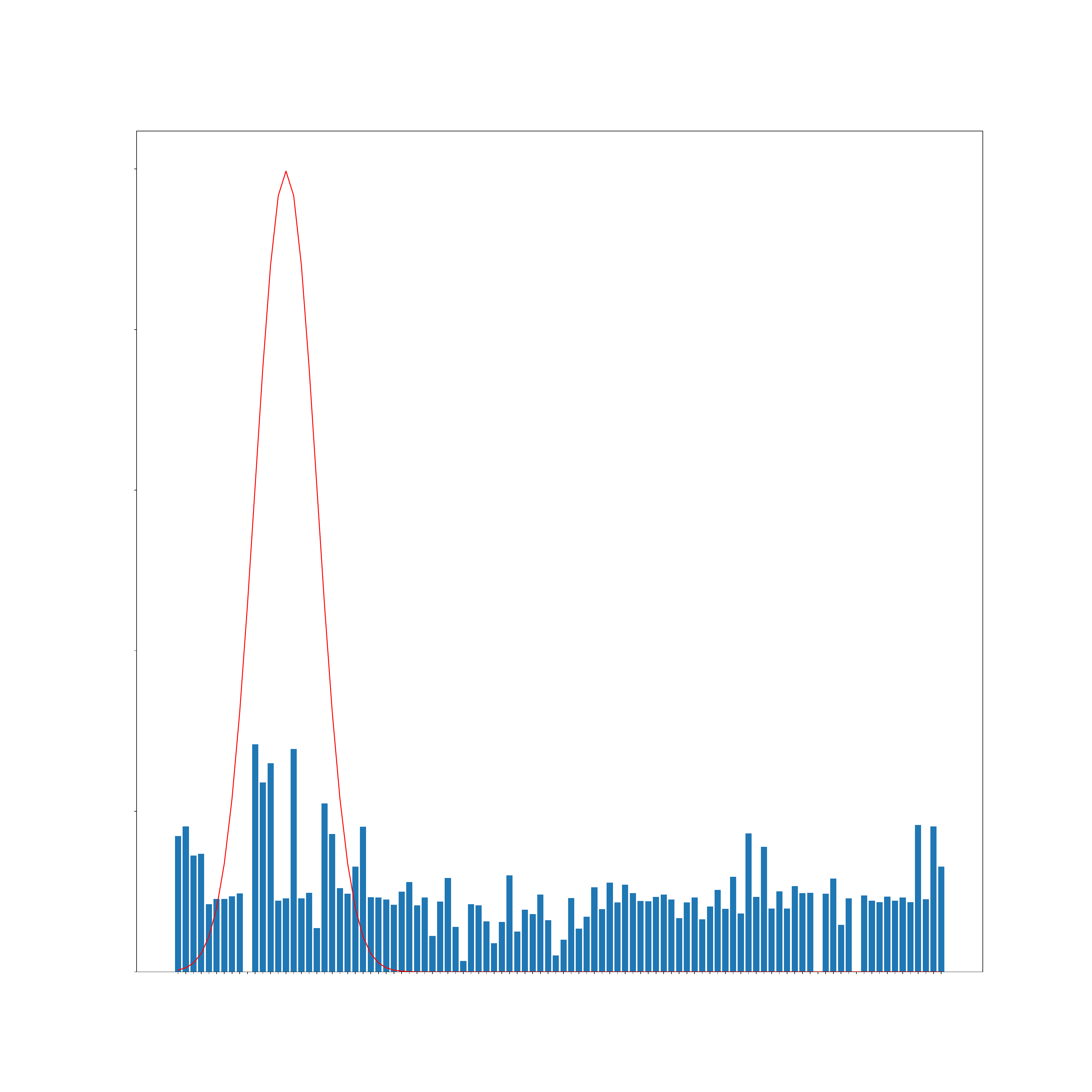}
\end{tabular} &

\begin{tabular}{@{}lr@{}}
\includegraphics[width=1.8cm,height=1.2cm,trim = 200 190 180 210,clip]{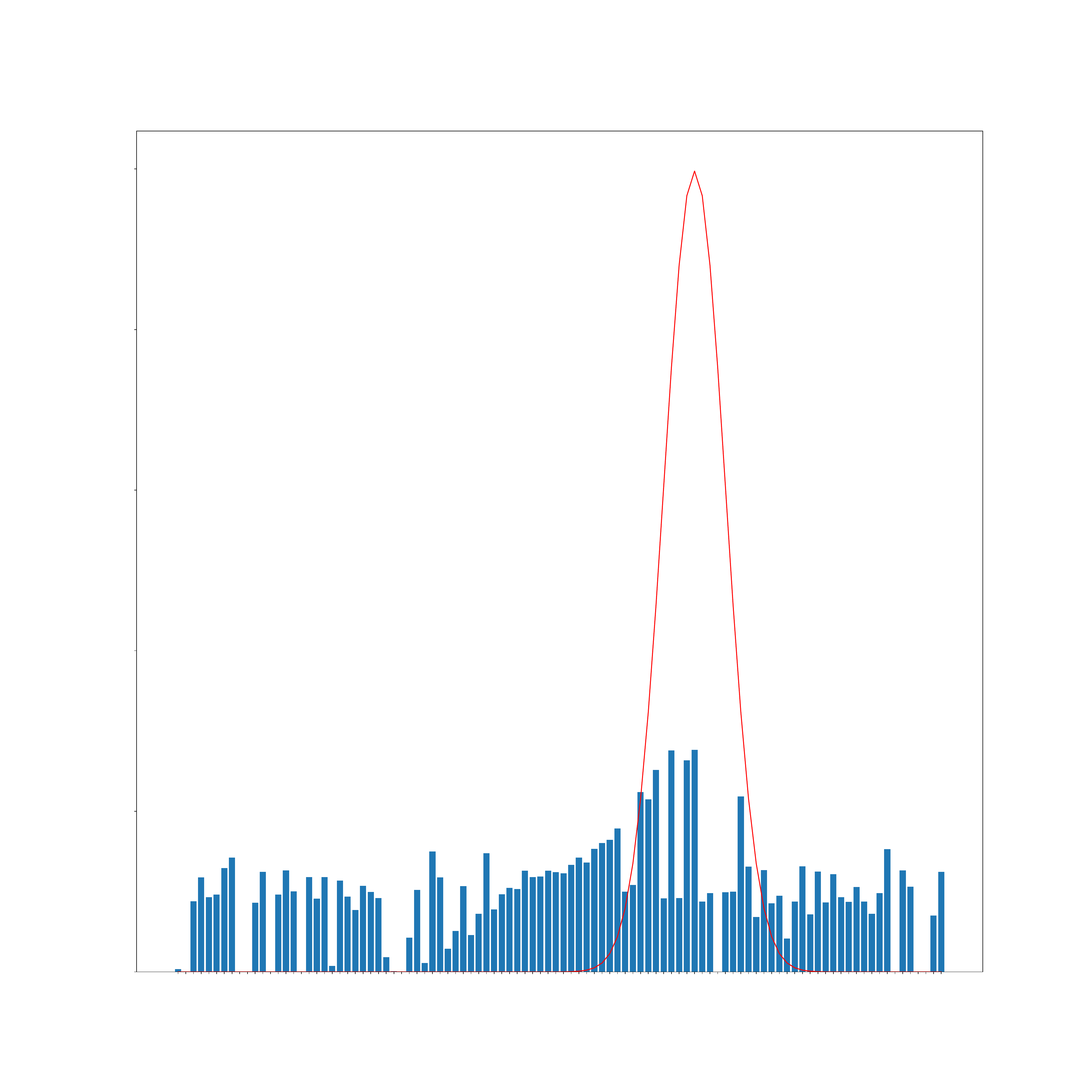} &
\includegraphics[width=1.8cm,height=1.2cm,trim = 200 190 180 210,clip]{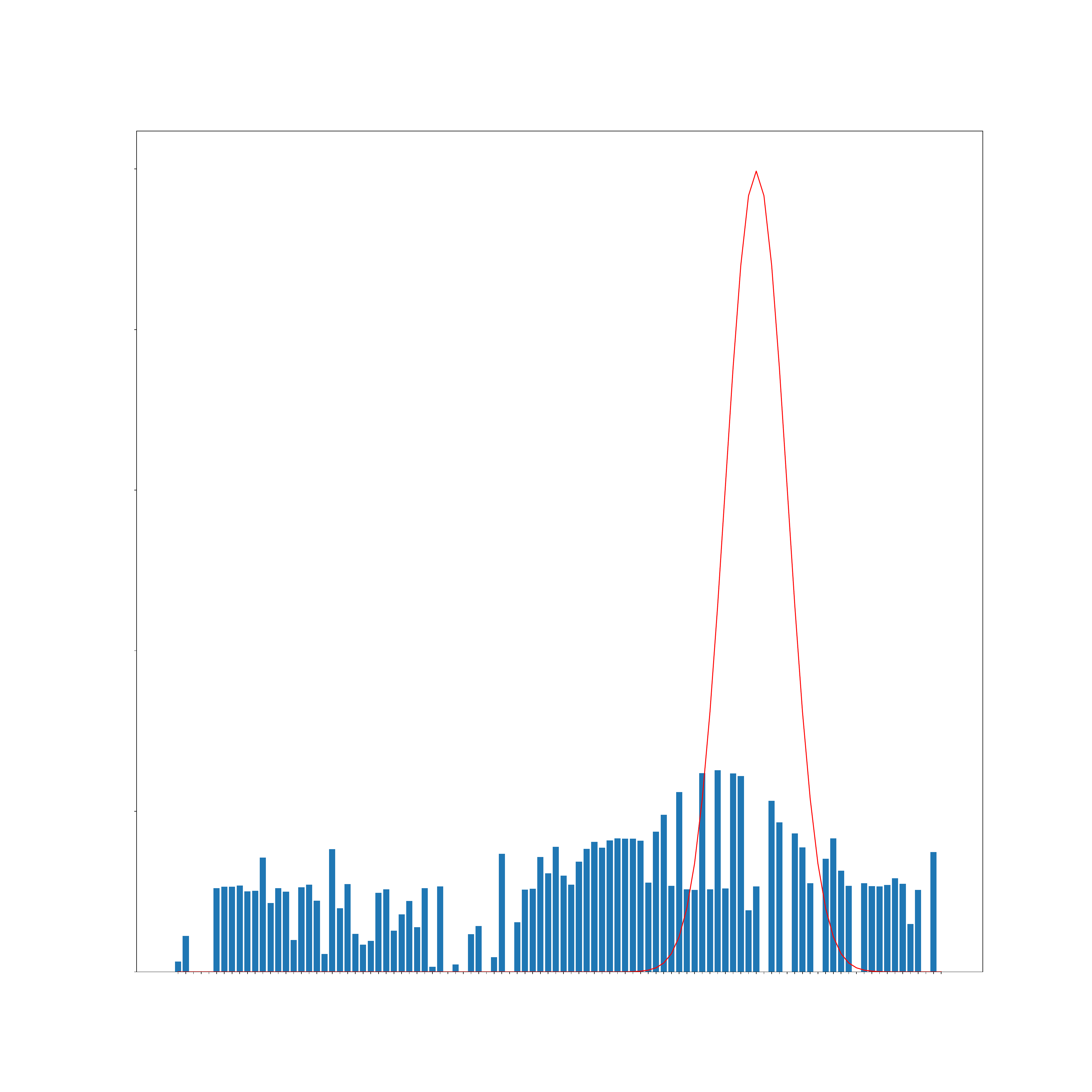} 
\end{tabular} &

\begin{tabular}{@{}lr@{}}
\includegraphics[width=1.8cm,height=1.2cm,trim = 200 190 180 210,clip]{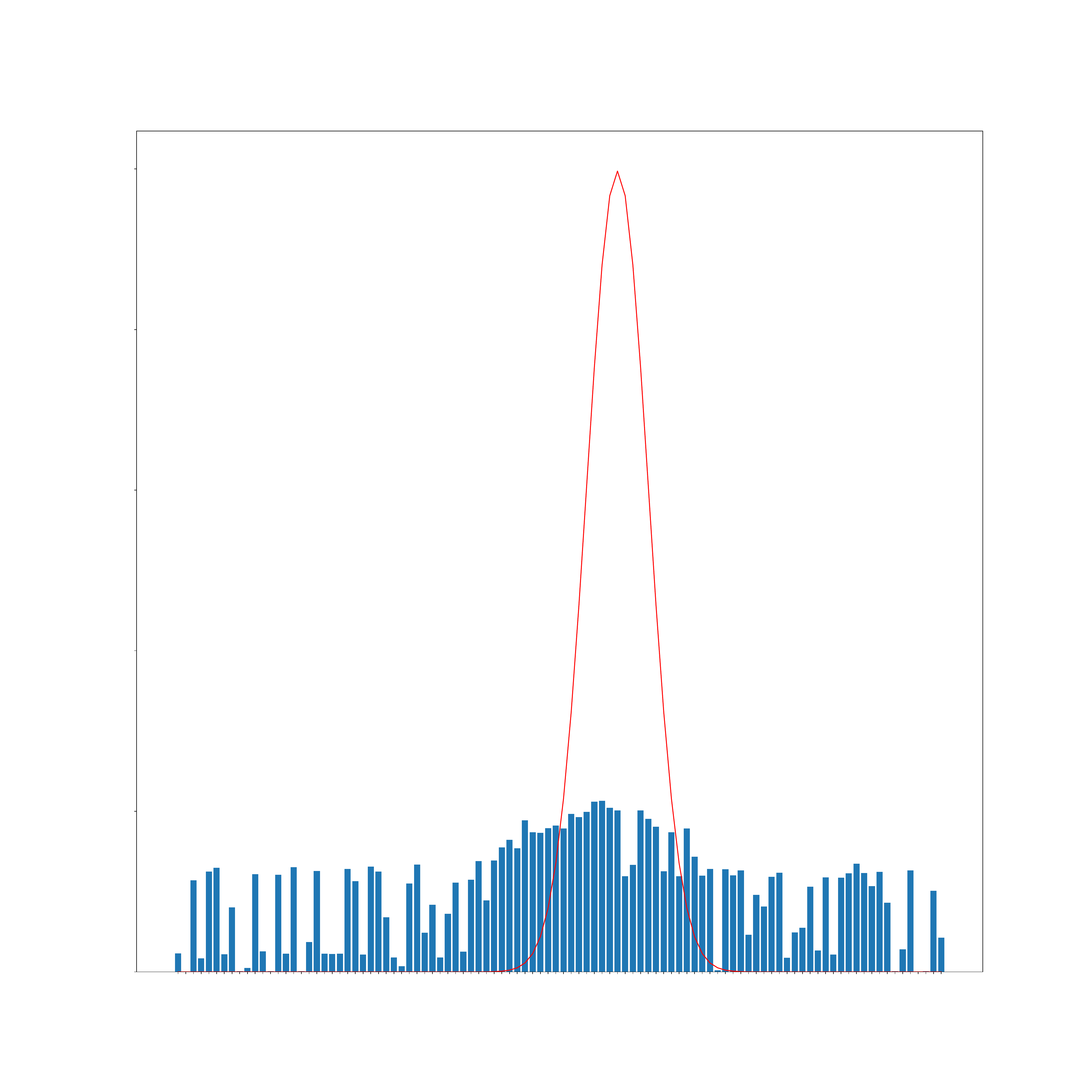} &
\includegraphics[width=1.8cm,height=1.2cm,trim = 200 190 180 210,clip]{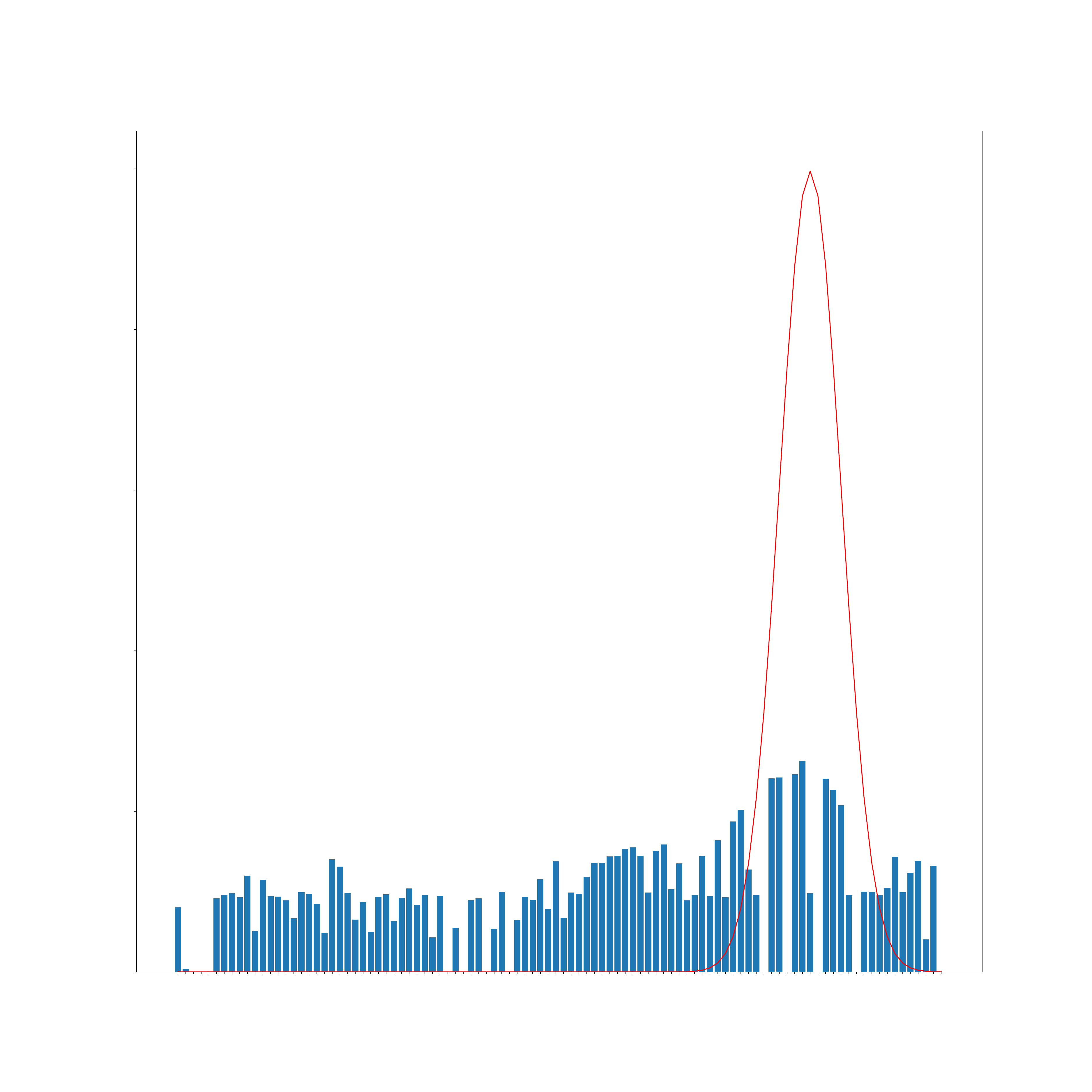}
\end{tabular}

\end{tabular}
\caption{Horizon line detection should be framed as a probabilistic regression problem due to its inherently stochastic nature. 
First row: Images where horizon line detection is easy (red line) and direct regression would work. 
Second row: Image where horizon line detection is hard and direct regression would not work. 
Plots below the images show the output probability distributions for the horizon line parameters $(\alpha,\rho)$, from the proposed method.
Red: ground truth; Blue: predicted density. Images are from the HLW dataset \cite{workman2016hlw}.}
	\label{fig:ambiguity_illustration}
    \vspace{-5mm}
\end{figure*}
{
    \small
    \bibliographystyle{ieeenat_fullname}
    \bibliography{main}

\begin{thebibliography}{35}
\providecommand{\natexlab}[1]{#1}
\providecommand{\url}[1]{\texttt{#1}}
\expandafter\ifx\csname urlstyle\endcsname\relax
  \providecommand{\doi}[1]{doi: #1}\else
  \providecommand{\doi}{doi: \begingroup \urlstyle{rm}\Url}\fi

\bibitem[Aodha et~al.(2013)Aodha, Humayun, Pollefeys, and Brostow]{aodha_tpami13}
Oisin~Mac Aodha, Ahmad Humayun, Marc Pollefeys, and Gabriel~J. Brostow.
\newblock Learning a confidence measure for optical flow.
\newblock \emph{{IEEE} Transactions on Pattern Recognition and Machine Intelligence ({TPAMI})}, 35:\penalty0 1107--1120, 2013.

\bibitem[Barinova et~al.(2010)Barinova, Lempitsky, Tretiak, and Kohli]{barinova2010geometric}
Olga Barinova, Victor Lempitsky, Elena Tretiak, and Pushmeet Kohli.
\newblock Geometric image parsing in man-made environments.
\newblock In \emph{Computer Vision--ECCV 2010: 11th European Conference on Computer Vision, Heraklion, Crete, Greece, September 5-11, 2010, Proceedings, Part II 11}, pages 57--70. Springer, 2010.

\bibitem[Brachmann and Rother(2019)]{brachmann2019neural}
Eric Brachmann and Carsten Rother.
\newblock Neural-guided {RANSAC}: Learning where to sample model hypotheses.
\newblock In \emph{Proceedings of the {IEEE/CVF} International Conference on Computer Vision}, pages 4322--4331, 2019.

\bibitem[Chang and Chen(2018)]{chang2018pyramid}
Jia-Ren Chang and Yong-Sheng Chen.
\newblock Pyramid stereo matching network.
\newblock In \emph{Proceedings of the IEEE conference on computer vision and pattern recognition}, pages 5410--5418, 2018.

\bibitem[Cheng et~al.(2008)Cheng, Wang, and Pollastri]{cheng2008neural}
Jianlin Cheng, Zheng Wang, and Gianluca Pollastri.
\newblock A neural network approach to ordinal regression.
\newblock In \emph{2008 IEEE international joint conference on neural networks (IEEE world congress on computational intelligence)}, pages 1279--1284. IEEE, 2008.

\bibitem[Forss\'en et~al.(2006)Forss\'en, Johansson, and Granlund]{fjg06}
Per-Erik Forss\'en, Bj\"orn Johansson, and G\"osta Granlund.
\newblock Channel associative networks for multiple valued mappings.
\newblock In \emph{2nd International Cognitive Vision Workshop}, pages 4--11, Graz, Austria, 2006.

\bibitem[Gal and Ghahramani(2016)]{gal2016dropout}
Yarin Gal and Zoubin Ghahramani.
\newblock Dropout as a bayesian approximation: Representing model uncertainty in deep learning.
\newblock In \emph{International Conference on Machine Learning}, pages 1050--1059. PMLR, 2016.

\bibitem[Garg et~al.(2020{\natexlab{a}})Garg, Wang, Hariharan, Campbell, Weinberger, and Chao]{div2020wstereo}
Divyansh Garg, Yan Wang, Bharath Hariharan, Mark Campbell, Kilian Weinberger, and Wei-Lun Chao.
\newblock Wasserstein distances for stereo disparity estimation.
\newblock In \emph{NeurIPS}, 2020{\natexlab{a}}.

\bibitem[Garg et~al.(2020{\natexlab{b}})Garg, Wang, Hariharan, Campbell, Weinberger, and Chao]{garg2020}
Divyansh Garg, Yan Wang, Bharath Hariharan, Mark Campbell, Kilian~Q. Weinberger, and Wei-Lun Chao.
\newblock Wasserstein distances for stereo disparity estimation.
\newblock In \emph{{NeurIPS}}, 2020{\natexlab{b}}.

\bibitem[Gneiting and Raftery(2007)]{gneiting07}
Tilmann Gneiting and Adrian~E. Raftery.
\newblock Strictly proper scoring rules, prediction, and estimation.
\newblock \emph{Journal of the American Statistical Association}, 102\penalty0 (477), 2007.

\bibitem[Gneiting et~al.(2007)Gneiting, Balabdaoui, and Raftery]{gneiting2007probabilistic}
Tilmann Gneiting, Fadoua Balabdaoui, and Adrian~E Raftery.
\newblock Probabilistic forecasts, calibration and sharpness.
\newblock \emph{Journal of the Royal Statistical Society Series B: Statistical Methodology}, 69\penalty0 (2):\penalty0 243--268, 2007.

\bibitem[H\"ager et~al.(2021)H\"ager, Persson, and Felsberg]{hager2021}
Gustav H\"ager, Mikael Persson, and Michael Felsberg.
\newblock Predicting disparity distributions.
\newblock In \emph{{IEEE} International Conference on Robotics and Automation ({ICRA'21})}, pages 4363--4369, 2021.

\bibitem[He et~al.(2016)He, Zhang, Ren, and Sun]{he2016residual}
Kaiming He, Xiangyu Zhang, Shaoqing Ren, and Jian Sun.
\newblock Deep residual learning for image recognition.
\newblock In \emph{{IEEE} Conference on Computer Vision and Pattern Recognition}, pages 770--778, 2016.

\bibitem[He et~al.(2019)He, Zhu, Wang, Savvides, and Zhang]{he2019bounding}
Yihui He, Chenchen Zhu, Jianren Wang, Marios Savvides, and Xiangyu Zhang.
\newblock Bounding box regression with uncertainty for accurate object detection.
\newblock In \emph{Proceedings of the ieee/cvf conference on computer vision and pattern recognition}, pages 2888--2897, 2019.

\bibitem[Hess et~al.(2022)Hess, Petersson, and Svensson]{hess2022object}
Georg Hess, Christoffer Petersson, and Lennart Svensson.
\newblock Object detection as probabilistic set prediction.
\newblock In \emph{17th European Conference on Computer Vision ({ECCV})}, pages 550--566. Springer, 2022.

\bibitem[Ilg et~al.(2018)Ilg, Cicek, Galesso, Klein, Makansi, Hutter, and Brox]{ilg2018uncertainty}
Eddy Ilg, Ozgun Cicek, Silvio Galesso, Aaron Klein, Osama Makansi, Frank Hutter, and Thomas Brox.
\newblock Uncertainty estimates and multi-hypotheses networks for optical flow.
\newblock In \emph{Proceedings of the European Conference on Computer Vision ({ECCV})}, pages 652--667, 2018.

\bibitem[Kendall and Gal(2017)]{kendall2017uncertainties}
Alex Kendall and Yarin Gal.
\newblock What uncertainties do we need in bayesian deep learning for computer vision?
\newblock \emph{Advances in neural information processing systems}, 30, 2017.

\bibitem[Kendall et~al.(2017)Kendall, Martirosyan, Dasgupta, Henry, Ryan, Bachrach, and Bry]{kendall2017stereo}
Alex Kendall, Hayk Martirosyan, Saumitro Dasgupta, Peter Henry, Ryan, Kennedy~Abraham Bachrach, and Adam Bry.
\newblock End-to-end learning of geometry and context for deep stereo regression.
\newblock In \emph{International Conference on Computer Vision ({ICCV})}, pages 66--75, 2017.

\bibitem[Kontschieder et~al.(2015)Kontschieder, Fiterau, Criminisi, and Bulo]{kontschieder2015deep}
Peter Kontschieder, Madalina Fiterau, Antonio Criminisi, and Samuel~Rota Bulo.
\newblock Deep neural decision forests.
\newblock In \emph{Proceedings of the IEEE international conference on computer vision}, pages 1467--1475, 2015.

\bibitem[Lakshminarayanan et~al.(2017)Lakshminarayanan, Pritzel, and Blundell]{lakshminarayanan2017simple}
Balaji Lakshminarayanan, Alexander Pritzel, and Charles Blundell.
\newblock Simple and scalable predictive uncertainty estimation using deep ensembles.
\newblock \emph{Advances in neural information processing systems}, 30, 2017.

\bibitem[Lee et~al.(2017)Lee, Kim, Lee, and Kim]{lee2017semantic}
Jun-Tae Lee, Han-Ul Kim, Chul Lee, and Chang-Su Kim.
\newblock Semantic line detection and its applications.
\newblock In \emph{Proceedings of the {IEEE} International Conference on Computer Vision}, pages 3229--3237, 2017.

\bibitem[Liu et~al.(2019)Liu, Zou, Che, Ding, Jia, You, and Kumar]{Liu_2019_ICCV}
Xiaofeng Liu, Yang Zou, Tong Che, Peng Ding, Ping Jia, Jane You, and B.V.K.~Vijaya Kumar.
\newblock Conservative wasserstein training for pose estimation.
\newblock In \emph{Proceedings of the IEEE/CVF International Conference on Computer Vision (ICCV)}, 2019.

\bibitem[Loh(2011)]{loh2011classification}
Wei-Yin Loh.
\newblock Classification and regression trees.
\newblock \emph{Wiley interdisciplinary reviews: data mining and knowledge discovery}, 1\penalty0 (1):\penalty0 14--23, 2011.

\bibitem[Mathieu et~al.(2016)Mathieu, Couprie, and LeCun]{mathieu2015deep}
Michael Mathieu, Camille Couprie, and Yann LeCun.
\newblock Deep multi-scale video prediction beyond mean square error.
\newblock In \emph{4th International Conference on Learning Representations, ICLR 2016}, 2016.

\bibitem[Mayer et~al.(2016)Mayer, Ilg, H{\"a}usser, Fischer, Cremers, Dosovitskiy, and Brox]{MIFDB16}
N. Mayer, E. Ilg, P. H{\"a}usser, P. Fischer, D. Cremers, A. Dosovitskiy, and T. Brox.
\newblock A large dataset to train convolutional networks for disparity, optical flow, and scene flow estimation.
\newblock In \emph{IEEE International Conference on Computer Vision and Pattern Recognition (CVPR)}, 2016.
\newblock arXiv:1512.02134.

\bibitem[Namdari and Li(2019)]{namdari2019review}
Alireza Namdari and Zhaojun Li.
\newblock A review of entropy measures for uncertainty quantification of stochastic processes.
\newblock \emph{Advances in Mechanical Engineering}, 11\penalty0 (6):\penalty0 1687814019857350, 2019.

\bibitem[Niu et~al.(2016)Niu, Zhou, Wang, Gao, and Hua]{niu2016ordinal}
Zhenxing Niu, Mo Zhou, Le Wang, Xinbo Gao, and Gang Hua.
\newblock Ordinal regression with multiple output cnn for age estimation.
\newblock In \emph{Proceedings of the IEEE conference on computer vision and pattern recognition}, pages 4920--4928, 2016.

\bibitem[Papernot and Mcdaniel(2018)]{Papernot2018DeepKN}
Nicolas Papernot and Patrick Mcdaniel.
\newblock Deep k-nearest neighbors: Towards confident, interpretable and robust deep learning.
\newblock \emph{ArXiv}, abs/1803.04765, 2018.

\bibitem[Rudnicki(2011)]{rudnicki2011shannon}
{\L}ukasz Rudnicki.
\newblock Shannon entropy as a measure of uncertainty in positions and momenta.
\newblock \emph{Journal of Russian Laser Research}, 32:\penalty0 393--399, 2011.

\bibitem[Schroff et~al.(2015)Schroff, Kalenichenko, and Philbin]{schroff2015}
Florian Schroff, Dmitry Kalenichenko, and James Philbin.
\newblock Facenet: A unified embedding for face recognition and clustering.
\newblock In \emph{Proceedings of the IEEE Conference on Computer Vision and Pattern Recognition (CVPR)}, pages 815--823, 2015.

\bibitem[Szegedy et~al.(2015)Szegedy, Liu, Jia, Sermanet, Reed, Anguelov, Erhan, Vanhoucke, and Rabinovich]{szegedy2015going}
Christian Szegedy, Wei Liu, Yangqing Jia, Pierre Sermanet, Scott Reed, Dragomir Anguelov, Dumitru Erhan, Vincent Vanhoucke, and Andrew Rabinovich.
\newblock Going deeper with convolutions.
\newblock In \emph{Proceedings of the IEEE conference on computer vision and pattern recognition}, pages 1--9, 2015.

\bibitem[Thorarinsdottir et~al.(2013)Thorarinsdottir, Gneiting, and Gissibl]{thorarinsdottir13}
Thordis~L. Thorarinsdottir, Tilmann Gneiting, and Nadine Gissibl.
\newblock Using proper divergence functions to evaluate climate models.
\newblock \emph{{SIAM/ASA} Journal on Uncertainty Quantification}, 1\penalty0 (1), 2013.

\bibitem[Thorpe(2018)]{thorpe18}
M. Thorpe.
\newblock Introduction to optimal transport.
\newblock Lecture Notes, 2018.

\bibitem[Workman et~al.(2016)Workman, Zhai, and Jacobs]{workman2016hlw}
Scott Workman, Menghua Zhai, and Nathan Jacobs.
\newblock Horizon lines in the wild.
\newblock In \emph{British Machine Vision Conference (BMVC)}, pages 20.1--20.12, 2016.
\newblock Acceptance rate: 39.4\%.

\bibitem[Zhang and O'Donnell(2020)]{ZHANG2020123}
Fan Zhang and Lauren~J. O'Donnell.
\newblock Chapter 7 - support vector regression.
\newblock In \emph{Machine Learning}, pages 123--140. Academic Press, 2020.

\end{thebibliography}
}


\end{document}